\documentclass{article} % For LaTeX2e
\usepackage{iclr2027_conference,times}

\usepackage{amsmath,amsfonts,bm}

\def\eqref#1{equation~\ref{#1}}
\def\1{\bm{1}}

\DeclareMathAlphabet{\mathsfit}{\encodingdefault}{\sfdefault}{m}{sl}
\SetMathAlphabet{\mathsfit}{bold}{\encodingdefault}{\sfdefault}{bx}{n}

\usepackage{hyperref}
\usepackage{url}
\usepackage{microtype}
\usepackage{graphicx}
\usepackage{subcaption}
\usepackage{booktabs}
\usepackage{enumitem}
\usepackage{float}
\usepackage{placeins}
\usepackage{amsmath}
\usepackage{amssymb}
\usepackage{mathtools}
\usepackage{amsthm}
\usepackage[ruled,vlined,linesnumbered]{algorithm2e}
\usepackage[capitalize,noabbrev]{cleveref}
\usepackage[textsize=tiny]{todonotes}

\theoremstyle{plain}

\theoremstyle{definition}

\theoremstyle{remark}

\title{Universe of Thoughts: A Computational Framework for Creative Reasoning in Large Language Models}

\author{
Yuto Suzuki \\
Department of Computer Science and Engineering \\
University of Colorado Denver \\
Denver, CO, USA \\
\texttt{yuto.suzuki@ucdenver.edu}
\And
Farnoush Banaei-Kashani \\
Department of Computer Science and Engineering \\
University of Colorado Denver \\
Denver, CO, USA \\
\texttt{farnoush.banaei-kashani@ucdenver.edu}
}

\iclrfinalcopy
\begin{document}

\maketitle
\lhead{Preprint}

\begin{abstract}
  Recent advances in Large Language Model (LLM) reasoning have improved conventional problem solving, but creative reasoning remains comparatively underexplored. Inspired by cognitive science, we formalize combinational, exploratory, and transformational creativity as executable computational operators over structured problem and solution spaces, specifying how each mode combines, explores, or transforms those spaces, respectively. Combinational reasoning transfers ideas across domains to form unfamiliar combinations; exploratory reasoning searches for new solutions within an existing conceptual space; and transformational reasoning modifies the rules or constraints that define that space. This formalization yields distinct algorithmic procedures, which we directly instantiate in Universe of Thoughts (UoT), an LLM reasoning framework. Existing creativity benchmarks emphasize either open-ended ideation or highly constrained problem solving. We therefore introduce three novel creative-reasoning tasks requiring concrete solutions in low-constraint settings. Across 10 generations per method and task, T-UoT with GPT-4o performs strongest on the low-constraint, high-objective-specificity Bridge and Electricity tasks, while C-UoT shows its strongest relative performance on the low-constraint, lower-objective-specificity Society task. In addition, we evaluate UoT on HypoArena, an independent scientific hypothesis-generation benchmark with 100 tasks across biomedical, machine-learning, and social-science domains. With Qwen3-14B, Exploratory UoT ranks first among seven reasoning methods, achieving a 32.7\% pairwise win rate compared with 25.5\% for the next-best method. Our results suggest distinct performance patterns across task structures: T-UoT is strongest in low-constraint, high-specificity settings, E-UoT in more constrained, high-specificity settings, and C-UoT in low-constraint, lower-specificity settings.

\end{abstract}

\section{Introduction}

Large Language Models (LLMs) have made rapid progress on complex reasoning,
supported by methods such as Chain-of-Thought (CoT) \citep{wei2022chain} and
Tree of Thoughts (ToT) \citep{yao2023tree}, which structure intermediate
reasoning and search \citep{achiam2023gpt,chowdhery2023palm}. These advances
have improved performance on problems with well-defined objectives and
verifiable solutions. Creative reasoning remains less explored, particularly for
open-ended or ill-defined problems where success requires solutions that are not
only valid but also novel and useful \citep{simon1973structure}. Such settings
include scientific hypothesis generation, drug discovery, and strategic
decision making, where progress may require moving beyond familiar solution
patterns \citep{singh2024transforming,gangwal2024generative}.

Recent work has begun to move LLM reasoning toward open-ended scientific discovery. Agentic systems such as The AI Scientist \citep{lu2024ai} automate broad scientific workflows through iterative idea generation, experimentation, and refinement, while methods such as MOOSE-Chem \citep{yang2024moose} more explicitly structure hypothesis generation around retrieving inspirations and composing them into new scientific ideas. Related work has further connected LLM generation to computational creativity, particularly through analogical or combinational mechanisms that bring together knowledge from different contexts \citep{gu2024llms,schapiro2025combinatorial}. These advances demonstrate that structured reasoning can support creative discovery. However, existing approaches typically instantiate a particular discovery workflow or a single creative mechanism; they do not provide a unified computational account of the qualitatively different ways in which creative reasoning can search for new solutions.

We address this gap with \emph{Universe of Thoughts} (UoT), a reasoning framework grounded in Margaret Boden's three modes of creativity: \emph{combinational}, \emph{exploratory}, and \emph{transformational} \citep{boden2004creative,boden2007creativity,boden2009computer}. UoT operationalizes these modes as distinct search programs: combinational reasoning transfers useful structure from unfamiliar contexts, exploratory reasoning expands into previously unvisited possibilities while preserving the current problem rules, and transformational reasoning modifies the assumptions that define which solutions can be considered. This unified formulation allows the three modes to be instantiated within the same LLM reasoning framework and, importantly, directly compared to characterize when different forms of creative search succeed or fail.

Evaluating creative reasoning remains difficult because existing benchmarks
often fall into two extremes. Open-ended idea-generation tasks permit novelty
but make utility difficult to assess, while grounded settings such as scientific
discovery provide clearer evaluation but impose stronger constraints that narrow
the solution space. We therefore characterize tasks by degree of constraint and
objective specificity. We introduce three low-constraint problem-solving tasks
that allow both novelty and utility to be evaluated, and additionally use
HypoArena \citep{zhong2026before}, which permits multiple plausible hypotheses
and pairwise comparison rather than requiring reproduction of a single reference
hypothesis. Together, these settings let us study how creative search behavior
changes with task structure.

The results reveal distinct performance patterns across task regimes. On the
novel creative-reasoning tasks, T-UoT performs strongest on the low-constraint,
high-objective-specificity Bridge and Electricity tasks, where flexible solution
spaces permit viable rule transformations and clear objectives support their
evaluation. C-UoT shows its strongest relative performance on Society, where
lower objective specificity broadens analogue retrieval and low constraint
facilitates cross-domain mechanism transfer. On HypoArena, E-UoT performs best
overall with Qwen3-14B, particularly in the more constrained, higher-specificity
biomedical and machine-learning settings, where local search is more strongly
guided by the problem context. Our failure
analysis localizes an additional bottleneck beyond creative search itself:
C-UoT often identifies useful analogies but fails to preserve their mechanisms
during final synthesis, while T-UoT can identify meaningful rule transformations
whose consequences are not fully carried into the final hypothesis. Targeted
interventions that preserve the transferred mechanism or explicitly reason
through the consequences of transformed rules substantially improve these
failure cases, suggesting that stronger downstream reasoning and integration
could allow these creative search processes to realize their potential more
reliably.

We summarize our main contributions as follows:

\begin{itemize}[leftmargin=*,topsep=2pt,itemsep=2pt,parsep=0pt,partopsep=0pt]
    \item We formalize and operationalize three cognitive-science creativity modes (combinational, exploratory, and transformational) as executable LLM search procedures in the Universe of Thoughts (UoT) framework.
    \item We evaluate UoT on novel creative-reasoning tasks in low-constraint settings
    and on 100 scientific hypothesis-generation tasks from HypoArena.
    \item We identify task-dependent strengths across creativity modes: transformational reasoning is strongest in low-constraint, high-objective-specificity settings, exploratory reasoning performs best in more constrained, higher-specificity settings, and combinational reasoning shows its strongest relative performance in low-constraint, lower-specificity settings. Ablation and failure analyses further reveal distinct contributions and bottlenecks across the three modes.
\end{itemize}
\section{Related Work}

\subsection{LLM Reasoning Frameworks}

LLM reasoning methods increasingly structure search over intermediate thoughts. Chain-of-Thought (CoT) elicits sequential reasoning \citep{wei2022chain}, Tree of Thoughts (ToT) introduces branching search with evaluation and backtracking \citep{yao2023tree}, and Graph of Thoughts (GoT) generalizes this process to arbitrary graph structures \citep{besta2024graph}. More recent methods enrich these search procedures through refinement, task-specific reasoning structures, self-reflection, and environment feedback, including Self-Discover \citep{zhou2024self} and Language Agent Tree Search \citep{zhou2023language}, as well as related refinement-oriented approaches \citep{shin2025enhancing}. Although such frameworks can support creative outputs, their mechanisms are primarily designed for generic reasoning, planning, and search rather than explicitly modeling distinct forms of creative reasoning. UoT instead operationalizes creativity-specific search procedures derived from cognitive science.

\subsection{Creative Reasoning in LLMs}

Prior work has explored individual mechanisms related to creative reasoning, including analogical reasoning \citep{yu2023thought,yasunaga2023large} and inductive reasoning \citep{cheng2024inductive,wang2023hypothesis}, as well as agentic systems for scientific discovery such as MOOSE-Chem, The AI Scientist, and AlphaEvolve \citep{yang2024moose,lu2024ai,novikov2025alphaevolve}. These systems show that iterative retrieval, recombination, refinement, and search can improve candidate solutions and sometimes produce novel outputs. However, they generally do not explicitly specify creativity-oriented search mechanisms for deliberately reaching unfamiliar solutions or changing the space of possibilities being explored.

More recent work explicitly connects LLMs with computational creativity, particularly combinational creativity \citep{schapiro2025combinatorial,gu2024llms}. In contrast, UoT provides a unified operationalization of combinational, exploratory, and transformational creativity as distinct search regimes: transferring structure across contexts, exploring new solutions within a fixed space, and modifying the rules that define the space itself. This common formulation enables direct comparison of the three modes and analysis of their distinct strengths and failure modes.

\subsection{Benchmarking Creativity in LLMs}
\label{sec:benchmarking-creativity}

Creativity benchmarks span a spectrum from open-ended generation to more constrained problem solving. Open-ended evaluations cover divergent thinking \citep{hou2025creativityprism}, story generation \citep{li2025automated}, scientific idea generation \citep{ruan2026evaluating,qiu2025ai,guo2025ideabench}, and creative writing \citep{sen2026automated}. These settings allow substantial novelty and diversity, but functional effectiveness can be difficult to assess objectively.

More constrained benchmarks provide clearer evaluation criteria, including MacGyver-style physical problem solving \citep{tian2024macgyver} and scientific-discovery tasks involving law discovery \citep{zheng2026newtonbench}, multimodal discovery \citep{liu2026towards,panigrahi2026heurekabench}, AI model discovery \citep{wu2026innovatorbench}, end-to-end research \citep{zhang2026innogym}, and scientific hypothesis generation \citep{liu2026researchbench,lew2026projectionbench,liu2025hypobench,chen2025hypospace,zhong2026before}. Their stronger grounding makes utility easier to evaluate, but also narrows the admissible solution space.

This leaves an important evaluation gap. Existing benchmarks often either
support open-ended idea generation, where utility is difficult to assess
concretely, or impose stronger constraints that narrow the admissible solution
space. We therefore introduce three low-constraint creative-reasoning tasks
that allow both novelty and utility to be evaluated while preserving multiple
valid solution strategies.

Because our primary contribution is the reasoning framework rather than
benchmark construction, we also evaluate UoT on HypoArena
\citep{zhong2026before}, a scientific hypothesis-generation benchmark that
permits multiple plausible hypotheses and pairwise comparison rather than
requiring reproduction of a single reference hypothesis.
\section{Universe of Thoughts: Creative Reasoning as Structured Search}
\label{sec:uot_methods}
\label{sec:computational_framework}

\begin{figure*}[t]
\centering
\includegraphics[width=\textwidth]{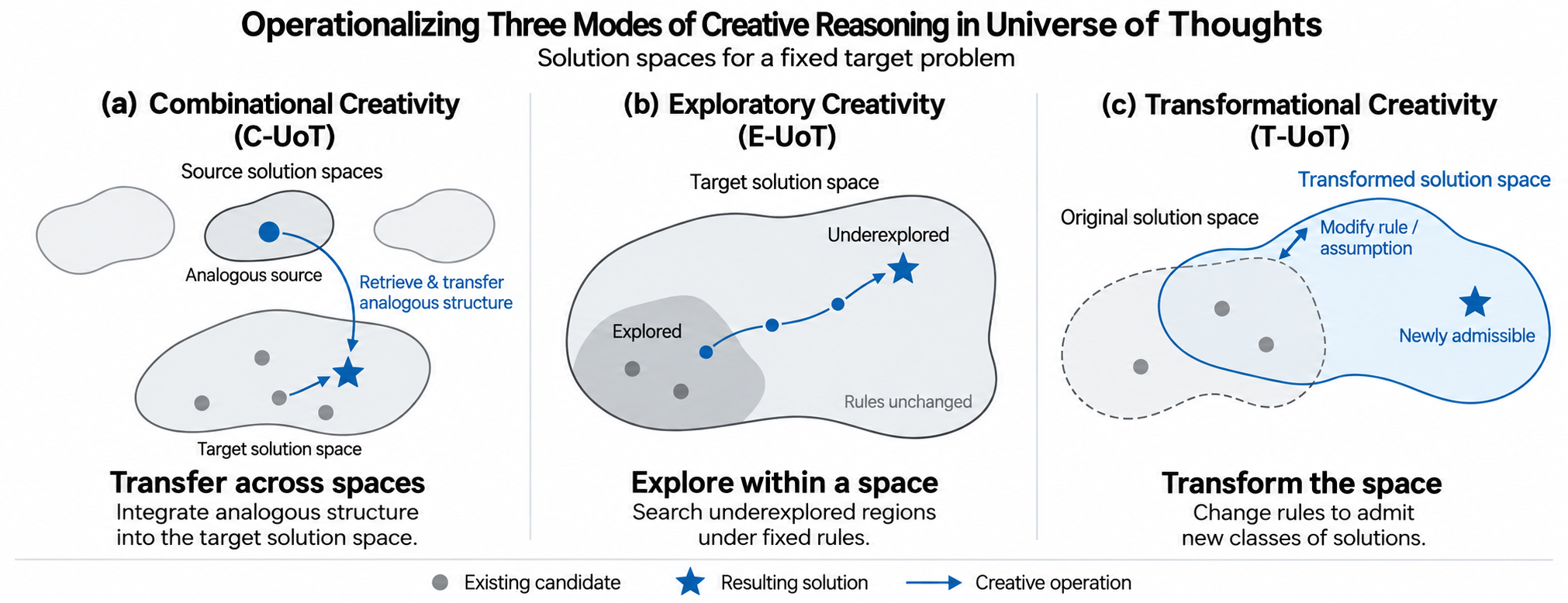}
\caption{\textit{
Three computational regimes of creative search. Combinational reasoning imports
transferable structure from a different context; exploratory reasoning expands from
existing candidates while preserving the current rules; and transformational reasoning
modifies the rules or assumptions that define the admissible solution space.
}}
\label{fig:reasoning_processes}
\end{figure*}

Margaret Boden distinguishes three forms of human creativity:
\emph{combinational}, \emph{exploratory}, and \emph{transformational}
\citep{boden2004creative,boden2007creativity,boden2009computer}.
We operationalize these modes as distinct computational search regimes that differ in
\emph{what the reasoning process changes} (Figure~\ref{fig:reasoning_processes}).
We follow Boden's original characterizations of the three modes; the source
definitions and their correspondence to our computational operators are
provided in Appendix~\ref{app:boden-definitions}.

Let $P$ denote a target problem, $H$ immutable task requirements, and $R$ mutable
rules or assumptions. The admissible solution space is
\[
\mathcal{S}(P,H,R)
=
\{s \mid s \text{ addresses } P \text{ while satisfying } H \text{ and } R\},
\]
and $\mathcal{K}_t \subseteq \mathcal{S}(P,H,R)$ denotes the solutions already explored.
Under this formulation, combinational reasoning changes the source structure used to
construct a solution, exploratory reasoning changes the candidate solution while keeping
$R$ fixed, and transformational reasoning changes $R$ itself.

\subsection{Combinational UoT}

Boden characterizes combinational creativity as producing
\emph{unfamiliar combinations of familiar ideas} by forming associations
between ideas that were previously only indirectly linked.
We interpret this computationally as transferring a useful structural
mechanism from a context that is functionally relevant to the target but
sufficiently unfamiliar to introduce a new combination.

Combinational UoT restructures direct generation through two intermediate
variables, an external analogue $a$ and an abstract mechanism $m$:
\[
\underbrace{p_{\theta}(s \mid P)}_{\text{direct search}}
\quad\Longrightarrow\quad
\underbrace{
p_{\theta}(a \mid P)\,
p_{\theta}(m \mid a)\,
p_{\theta}(s \mid P,m)
}_{\text{combinational search}}.
\]
The product denotes a sequence of conditional generation stages rather than an
exact factorization of $p_{\theta}(s \mid P)$.

The analogue is therefore selected to be both \emph{functionally relevant}
to the target and \emph{contextually unfamiliar} relative to previously
explored solutions. This differs from Analogical Prompting
\citep{yasunaga2023large}, which primarily self-generates relevant examples
or knowledge to improve downstream reasoning. C-UoT differs both in
\emph{objective} and \emph{structure}: it deliberately searches for analogues
from unfamiliar contexts to promote creative recombination, and explicitly
decomposes analogical reasoning into
\emph{analogue $\rightarrow$ mechanism abstraction $\rightarrow$ target transfer}.

Operationally, C-UoT proposes distant analogues, solves the source problem,
abstracts the transferable mechanism, maps that mechanism into the target
context, and evaluates the resulting candidate. This explicit decomposition
also makes the reasoning process inspectable, allowing failures to be
localized to analogue retrieval, mechanism abstraction, or downstream
transfer.

\subsection{Exploratory UoT}

Boden characterizes exploratory creativity as searching within an existing
conceptual space to discover previously unvisited possibilities without changing
the rules that define that space. We interpret this computationally as expanding
the set of candidate solutions while preserving the current problem formulation
and mutable rules.

Exploratory UoT structures search around an already explored solution
$s_k \in \mathcal{K}_t$ while preserving the current rules:
\[
\underbrace{p_{\theta}(s \mid P)}_{\text{direct search}}
\quad\Longrightarrow\quad
\underbrace{
p_{\theta}(s_k \mid P,\mathcal{K}_t)\,
p_{\theta}(s' \mid P,s_k,R)
}_{\text{exploratory search}},
\qquad R \text{ fixed}.
\]
The product denotes a sequence of conditional generation stages rather than an
exact factorization of $p_{\theta}(s \mid P)$.

Directly searching for novel solutions in a large conceptual space is difficult.
E-UoT therefore operationalizes exploration locally: rather than generating an
entirely new solution from scratch, it starts from an existing candidate and
searches for neighboring alternatives that have not yet been explored. Repeating
this process expands the frontier of $\mathcal{K}_t$ and gradually reaches new
regions of the current solution space.

Operationally, E-UoT iteratively proposes such neighboring candidates, evaluates
them, and retains promising ones for further expansion. The defining constraint is
that the problem formulation and governing rules remain fixed throughout the
search, so novelty arises from traversing previously unvisited parts of the existing
space rather than importing external structure or transforming the space itself.

\subsection{Transformational UoT}

Boden characterizes transformational creativity as changing the defining
dimensions of a conceptual space so that possibilities excluded by the
original space can become reachable. We interpret this computationally as
modifying selected mutable rules or assumptions $R$ while preserving the
underlying problem and its immutable requirements $H$.

Transformational UoT searches over alternative rule sets $R'$ while preserving
the immutable task requirements $H$:
\[
\underbrace{p_{\theta}(s \mid P,H,R)}_{\text{direct search}}
\quad\Longrightarrow\quad
\underbrace{
p_{\theta}(R' \mid P,H,R)\,
p_{\theta}(s \mid P,H,R')
}_{\text{transformational search}}.
\]
The product denotes a sequence of conditional generation stages rather than an
exact factorization of $p_{\theta}(s \mid P,H,R)$.

Not every change to $R$ constitutes a useful transformation. T-UoT therefore
retains only candidate rule changes that preserve the original task and its
core difficulty while opening a meaningfully different region of the solution
space. 

\subsection{Three Executable Creative Search Programs}

UoT does not claim analogy, iterative search, or rule modification as new
primitives. Rather, it operationalizes an established theory of creativity as
three executable and empirically testable search programs. Full
pseudocode, operator definitions, selection procedures, and hyperparameters
are provided in Appendix~\ref{sec:uot-algorithms} and
Appendix~\ref{sec:uot-scoring-details}.
\section{Experimental Setup}
\label{sec:experiments}

We evaluate UoT across task regimes characterized by two properties that directly shape the creative search problem: \emph{degree of constraint} and \emph{objective specificity}. The degree of constraint captures how strongly problem requirements restrict the feasible solution space; stronger constraints narrow the set of admissible solutions and limit the extent to which alternative mechanisms, rules, or solutions can be accommodated. Objective specificity captures how precisely the desired outcome is defined. A more specific objective narrows the search target and provides a clearer criterion for evaluating candidate solutions, whereas a less specific objective permits broader notions of relevance and similarity, expanding the range of potentially useful analogies that can be retrieved from other domains. These dimensions therefore affect both the space over which creative reasoning operates and how that space can be searched and evaluated.

As illustrated in Fig.~\ref{fig:task-uot-categorization}, we evaluate UoT in two complementary settings spanning these dimensions. HypoArena \citep{zhong2026before} provides externally defined scientific hypothesis-generation tasks with relatively high degrees of constraint, while spanning higher- and lower-specificity objectives. To complement this regime, we introduce three novel creative-reasoning tasks with substantially lower degrees of constraint and expansive solution spaces: Bridge and Electricity provide relatively specific objectives, while Society evaluates reasoning under a less specific objective. Together, these settings cover distinct regions of the task space and allow us to examine how creative reasoning methods behave as the structure of the underlying search problem changes.

\subsection{Novel Creative-Reasoning Tasks}

We construct three tasks spanning transportation, energy, and social
intervention: \textit{One-Lane Bridge}, which requires designing a
traffic-control policy under fixed crossing constraints; \textit{Electricity
Tariff}, which requires reducing peak demand without curtailing critical
usage; and \textit{Social Cohesion}, which requires increasing cross-group
interaction while preserving privacy, inclusivity, and safety. Full task
definitions are provided in Appendix~\ref{sec:controlled-tasks}.

Following the standard view of creativity as requiring both novelty and
usefulness \citep{runco2012standard}, we report \textit{Creativity} as the
mean of Novelty and Utility, with Feasibility reported separately as a
validity measure. Outputs are canonicalized before evaluation to reduce
effects of response length and presentation. Ten PhD students provide human
judgments, and Gemini~2.5~Pro provides automated evaluations. Full scoring
procedures, novelty computation, prompts, and known-solution baselines are provided
in Appendix~\ref{sec:evaluation-metrics-appendix}.

\subsection{HypoArena}

We evaluate on a fixed panel of 100 HypoArena tasks from its three
single-hypothesis research domains: biomedical science, machine learning, and
social science (34/34/32 tasks). Final outputs are canonicalized into their
core hypotheses before pairwise evaluation.

We follow the HypoArena evaluation protocol using Gemma-3-12B as the primary
judge. Pairwise judgments are aggregated with Bradley--Terry--Davidson (BTD)
ratings, and we additionally report pairwise win rate. HypoArena emphasizes
contextual grounding and evidential justification while also rewarding
non-obvious inferential insight, making it suitable for evaluating creative
reasoning under scientific constraints. Additional evaluation details and
judge-robustness analyses are provided in the Appendix
(Appendix~\ref{sec:hypoarena-protocol} and
Appendix~\ref{sec:judge-consistency}).

\subsection{Baselines and Backbones}

We compare UoT against general-purpose reasoning baselines spanning direct
generation and increasingly structured search: OnePrompt, Chain-of-Thought
(CoT) \citep{wei2022chain}, Tree of Thoughts (ToT) \citep{yao2023tree}, and
Enhanced Graph of Thoughts (EGoT) \citep{shin2025enhancing}. We focus on
matched general-purpose reasoning frameworks rather than specialized
scientific-discovery systems, which often incorporate additional retrieval
pipelines, tools, or domain-specific assumptions.

For the novel creative-reasoning tasks, GPT-4o is used as the shared backbone for UoT and all
matched reasoning baselines; GPT-5 and DeepSeek V3.1 are included only as
contextual reference models. For HypoArena, Qwen3-14B is the primary shared
backbone, and GLM-4-9B is additionally evaluated to assess backbone
sensitivity. Full implementation settings are provided in
Appendix~\ref{sec:implementation-details}.

\begin{figure*}[t]
\centering
\includegraphics[width=\textwidth]{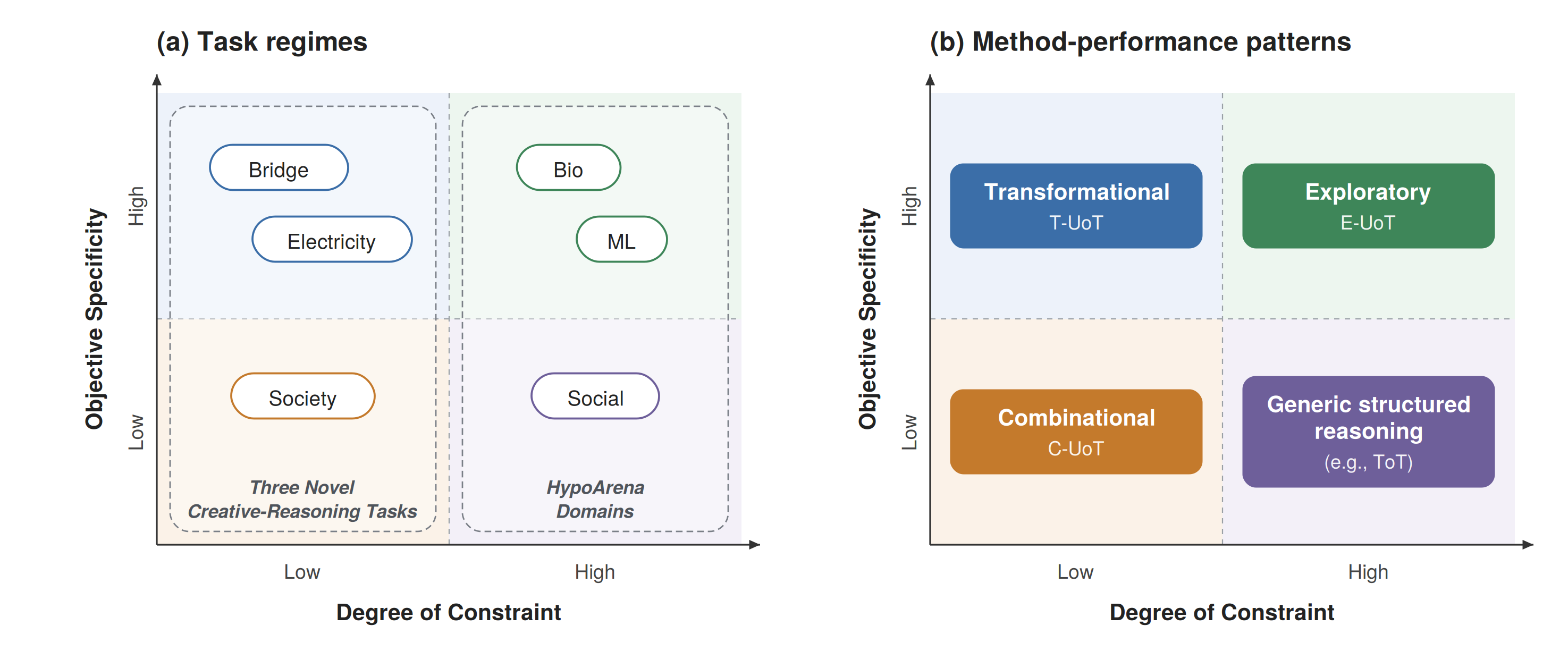}
\caption{Task regimes and method-performance patterns along degree of constraint
and objective specificity. (a) Bridge and Electricity are low-constraint,
high-specificity design tasks, while Society is low-constraint and lower-specificity.
HypoArena Bio and ML are high-constraint, high-specificity hypothesis-generation
tasks, while HypoArena Social is high-constraint and lower-specificity.
(b) The observed performance patterns are consistent with the search mechanisms
of T-UoT, C-UoT, and E-UoT, while the high-constraint, low-specificity regime
shows stronger performance from generic structured reasoning methods such as ToT.}
\label{fig:task-uot-categorization}
\end{figure*}

\section{Results}
\label{sec:results}

\subsection{Novel Creative-Reasoning Task Results}

\begin{table*}[t]
  \caption{Novel creative-reasoning task creativity results. Human and LLM-based creativity
  scores across the three novel creative-reasoning tasks. Creativity is the mean of Novelty
  and Utility. Full Novelty, Utility, and Feasibility results are reported in
  Appendix~\ref{sec:controlled-full-results}. Bold indicates the best result
  among matched GPT-4o reasoning methods; underlined values indicate the best
  contextual reference model.}
  \label{tab:controlled-creativity}
  \centering
  \small
  \begin{tabular}{l cccccc}
  \toprule
  & \multicolumn{2}{c}{\bf Bridge}
  & \multicolumn{2}{c}{\bf Electricity}
  & \multicolumn{2}{c}{\bf Society} \\
  \cmidrule(lr){2-3} \cmidrule(lr){4-5} \cmidrule(lr){6-7}
  \bf Method
  & \bf Human C & \bf LLM C
  & \bf Human C & \bf LLM C
  & \bf Human C & \bf LLM C \\
  \midrule
  OnePrompt     & 0.480 & 0.566 & 0.540 & 0.508 & 0.550 & 0.522 \\
  CoT           & 0.590 & 0.651 & 0.560 & 0.450 & 0.500 & 0.586 \\
  ToT           & 0.580 & 0.579 & 0.530 & 0.488 & 0.530 & 0.574 \\
  EGoT          & 0.650 & 0.641 & 0.500 & 0.409 & \textbf{0.560} & 0.613 \\
  C-UoT         & 0.580 & 0.664 & 0.440 & 0.512 & \textbf{0.560} & \textbf{0.649} \\
  E-UoT         & 0.600 & 0.589 & 0.430 & 0.463 & 0.550 & 0.545 \\
  T-UoT         & \textbf{0.680} & \textbf{0.698}
                & \textbf{0.620} & \textbf{0.526}
                & 0.550 & 0.628 \\
  \midrule
  GPT-5         & \underline{0.620} & \underline{0.649}
                & \underline{0.580} & 0.504
                & 0.640 & 0.631 \\
  DeepSeek V3.1 & 0.460 & 0.458
                & \underline{0.580} & \underline{0.571}
                & \underline{0.720} & \underline{0.723} \\
  \bottomrule
  \end{tabular}
  \end{table*}

  Table~\ref{tab:controlled-creativity} summarizes Creativity scores under human and automated evaluation. Among matched GPT-4o reasoning methods, T-UoT performs strongest on Bridge and Electricity, achieving the highest human-rated Creativity on Bridge ($0.680$) and Electricity ($0.620$), as well as the highest LLM-rated scores on Bridge ($0.698$) and Electricity ($0.526$).

As shown in Fig.~\ref{fig:task-uot-categorization}, Bridge and Electricity combine relatively low degrees of constraint with high objective specificity. A lower degree of constraint leaves greater flexibility in the rule-generation stage $p_{\theta}(R' \mid P,H,R)$, since more alternative rule sets $R'$ can modify the search space while still permitting solutions that satisfy the immutable requirements $H$. High objective specificity, in turn, provides a clearer criterion for evaluating and selecting among the generated rule sets $R'$, helping T-UoT identify transformations that are most promising for the target objective.

C-UoT shows the complementary pattern on Society, where it ties for the highest human-rated Creativity among matched GPT-4o methods ($0.560$) and achieves the highest LLM-rated score ($0.649$). As illustrated in Fig.~\ref{fig:task-uot-categorization}, Society combines a low degree of constraint with relatively low objective specificity. Lower objective specificity makes the analogue-retrieval stage $p_\theta(a\mid P)$ less restrictive, allowing a broader range of problems from other domains to be considered functionally relevant to the target. Once a transferable mechanism $m$ is identified, the low degree of constraint makes the transfer stage $p_\theta(s\mid P,m)$ less restrictive, since fewer hard requirements must be satisfied when adapting that mechanism to the target problem. Together, these conditions favor both analogue discovery and cross-domain mechanism transfer, consistent with C-UoT's stronger relative performance on Society.

The full
Novelty, Utility, and Feasibility breakdown is provided in
Appendix~\ref{sec:controlled-full-results}. Human and automated evaluations show
substantial agreement at the method level ($r{=}0.708$;
Table~\ref{tab:human-llm-corr}), supporting the use of automated scores as a
complement to human judgments. Representative examples in Appendix~\ref{app:bridge-examples} illustrate the
distinct search behaviors induced by the three UoT modes.

\subsection{HypoArena Results}

On \texttt{Qwen3-14B} (Table~\ref{tab:hypoarena-qwen-100}), E-UoT achieves
the strongest overall performance, ranking first in both biomedical science and
machine learning, while ToT performs best in social science. E-UoT reaches an
average BTD rating of 1535.4 and a pairwise win rate of 32.7\%, compared with
1515.8 and 25.5\% for the next-best method, ToT. With GLM-4-9B, OnePrompt
ranks first while E-UoT remains the strongest structured method, indicating
backbone dependence (Table~\ref{tab:hypoarena-glm-100}).

As illustrated in Fig.~\ref{fig:task-uot-categorization}, HypoArena tasks have a relatively high degree of constraint, while the biomedical and machine-learning domains additionally provide relatively specific objectives. Both properties make the local expansion stage $p_{\theta}(s' \mid P,s_k,R)$ more directed. A higher degree of constraint narrows the admissible solution space, reducing the range of viable modifications around the current candidate $s_k$. High objective specificity provides additional problem context about what constitutes a useful improvement, further guiding the search toward promising neighboring solutions. These conditions are therefore well matched to E-UoT's strategy of iteratively expanding within the existing solution space while keeping the governing rules $R$ fixed. 

C-UoT and T-UoT are less reliable on average in these more constrained regimes,
as cross-domain transfer and rule transformation must satisfy a narrower set of
admissible conditions. Nevertheless, their lower aggregate rankings do not imply
that these search modes are uniformly unsuccessful. Under the primary Gemma
judge, C-UoT is top-ranked on 15/100 HypoArena tasks (8 uniquely), while T-UoT
is top-ranked on 11/100 tasks (6 uniquely). These task-specific advantages also
recur across judges: C-UoT and T-UoT outperform ToT on 26 and 24 tasks,
respectively, under at least two of three judges, and on 11 tasks each under all
three judges (Appendix~\ref{sec:hypoarena-qualitative}). Thus, although
exploratory search is more reliable on average in the highly constrained
HypoArena setting, combinational and transformational search can still uncover
stronger hypothesis directions on particular problems.

The main ranking conclusions are not specific to the Gemma judge:
re-scoring the same frozen Qwen3-14B outputs with two additional judges
preserves the relative advantages of E-UoT over OnePrompt and ToT
($\rho=0.59$--$0.79$ across judge pairs), while hard preference reversals
remain rare (2.6--5.4\%; Appendix~\ref{sec:judge-consistency}).

\begin{table}[t]
\caption{Qwen3-14B on HypoArena (100 tasks; Gemma-3-12B judge; canonicalized core hypotheses). Bio\,34 / ML\,34 / Social\,32. Bio/ML/Social/Avg: pooled BTD ratings centered at 1500. WR: pairwise win rate (\%). Bold: column best.}
\label{tab:hypoarena-qwen-100}
\centering
\small
\begin{tabular}{r l ccccc}
\toprule
\bf Rank & \bf Method & \bf Bio & \bf ML & \bf Social & \bf Avg & \bf WR (\%) \\
\midrule
1 & E-UoT     & \textbf{1541.8} & \textbf{1547.1} & 1516.3          & \textbf{1535.4} & \textbf{32.7} \\
2 & ToT       & 1513.2          & 1510.4          & \textbf{1524.5} & 1515.8          & 25.5 \\
3 & EGoT      & 1482.7          & 1510.6          & 1521.1          & 1505.1          & 25.3 \\
4 & CoT       & 1486.2          & 1493.3          & 1501.9          & 1493.7          & 18.8 \\
5 & C-UoT     & 1495.1          & 1477.6          & 1497.0          & 1489.6          & 19.5 \\
6 & T-UoT     & 1490.1          & 1480.0          & 1484.6          & 1484.6          & 17.3 \\
7 & OnePrompt & 1490.9          & 1481.0          & 1454.5          & 1475.7          & 14.3 \\
\bottomrule
\end{tabular}
\end{table}

\paragraph{Operator ablation.}
To isolate individual creativity-specific operations, we remove the structured
cross-domain transfer in C-UoT, the option to switch between mechanisms during
E-UoT expansion, and rule transformation in T-UoT while otherwise preserving
each mode's search procedure. A position-debiased pairwise judge compares full
and ablated outputs on the same 100 HypoArena tasks
(Table~\ref{tab:uot-ablation-domain}).

\begin{table}[t]
\caption{Operator ablation on 100 HypoArena tasks. Values are mean pairwise
win rates of the full variant against its ablated counterpart
(chance $=50\%$).}
\label{tab:uot-ablation-domain}
\centering
\small
\begin{tabular}{lcccc}
\toprule
\bf Mode & \bf Bio & \bf ML & \bf Social & \bf All \\
\midrule
C-UoT & \textbf{58.8\%} & \textbf{57.0\%} & \textbf{59.0\%} & \textbf{58.3\%} \\
E-UoT & 53.3\% & 49.2\% & 55.0\% & 52.4\% \\
T-UoT & 54.1\% & 53.4\% & 55.1\% & 54.2\% \\
\bottomrule
\end{tabular}
\end{table}

The largest ablation effect occurs for C-UoT, whose full variant is preferred in
58.3\% of comparisons, showing that structured cross-domain transfer contributes
beyond the surrounding pipeline. T-UoT is preferred in 54.2\% of comparisons,
providing a smaller but consistent indication that explicitly modifying the rule
set can expose useful hypothesis directions beyond the same downstream search
without transformation. For E-UoT, the ablation removes only mechanism switching while retaining the
same ToT-style search controller and local refinement of the current hypothesis.
Full E-UoT wins 52.4\% of these comparisons, indicating only a modest incremental
benefit from mechanism switching. Together with E-UoT's substantially stronger
performance than ToT in Table~\ref{tab:hypoarena-qwen-100}, this suggests that
its overall advantage is not driven primarily by mechanism switching alone, but
by the broader exploratory search procedure, including local refinement.

\paragraph{Failure-mode interventions.}
Stage-level traces suggest that C-UoT often loses retrieved mechanisms during
final synthesis, while T-UoT may fail to propagate the consequences of selected
rule transformations (Appendix~\ref{sec:uot-intervention}). On frozen failure
cases, preserving the C-UoT analogy mapping improves average rank from
$\#4$ to $\#1.5$, while explicitly deriving a mechanistic consequence from the
selected T-UoT transformation improves rank from $\#7$ to $\#2.5$.
These diagnostic interventions suggest that useful creative structure can be
discovered upstream but lost during final synthesis; full intervention details
are provided in Appendix~\ref{sec:uot-intervention}. This suggests that the limitation lies partly in the backbone's ability to
reason over and integrate the discovered creative structure, and that stronger
reasoning capabilities could allow these search modes to realize their benefits
more reliably.

\section{Conclusions and Future Work}

We introduced Universe of Thoughts (UoT), a computational framework that operationalizes combinational, exploratory, and transformational creativity as distinct executable search processes for Large Language Models. Rather than treating creativity as a single undifferentiated capability, UoT separates cross-domain recombination, within-space exploration, and transformation of the rules that define the solution space.

Across complementary creative-reasoning and scientific hypothesis-generation settings, the three search modes exhibit distinct performance patterns depending on task structure. Combinational search performs particularly well when broader objectives permit wider analogue discovery and relatively few constraints facilitate mechanism transfer; exploratory search benefits from more constrained, higher-specificity settings that provide stronger guidance for local improvement; and transformational search performs strongly when flexible solution spaces permit viable rule changes while clear objectives support their evaluation. These results suggest that the effectiveness of different creative search processes depends in part on the structure of the problem, rather than one mode being uniformly strongest across settings.

The current implementation nevertheless depends substantially on the reasoning, search, abstraction, and transfer capabilities of the underlying foundation model. Future work can therefore investigate models trained explicitly to support these creative operations, with the goal of improving their reliability and effectiveness across task regimes. Such models could ultimately be integrated with retrieval, experimentation, and feedback loops toward more capable systems for autonomous scientific discovery.

% ICLR 2027 required / recommended end-matter (before references).
% These do not count toward the main-text page limit. Fill in before submission.

\section*{AI Use Statement}

We used generative AI tools to assist with software implementation and debugging, data analysis and interpretation, literature review, preparation and refinement of scientific figures, and feedback on experimental methodology and presentation. Generative AI was also used to edit, reorganize, and refine portions of the
manuscript for clarity, structure, grammar, and readability, based on research
content, interpretations, and conclusions developed by the authors.

Generative AI was not used to originate the core research question or the central conceptual framework of this work, nor to independently determine the scientific conclusions of the study. Research decisions, experimental configurations, interpretation of the final results, and all claims presented in the paper were reviewed and determined by the authors. AI-generated or AI-assisted code was tested and verified by the authors, and AI-assisted literature discussions and manuscript content were checked against the cited sources.

The authors reviewed all AI-assisted work and take full responsibility for the final content of the paper, including its methods, experiments, analyses, claims, text, code, and figures.

\section*{Ethics Statement}

This work studies computational methods for creative reasoning with large language models. The experiments do not involve deployment of generated hypotheses or solutions in real-world scientific, medical, or other high-stakes settings. Generated scientific hypotheses are evaluated only as research artifacts and should not be interpreted as experimentally validated scientific conclusions or recommendations.

Our evaluation uses both human and LLM-based judgments. Human evaluators assess model outputs rather than participating as subjects of behavioral or clinical experimentation. As with any subjective evaluation of creativity, judgments of novelty, usefulness, and hypothesis quality may reflect evaluator preferences or biases. We therefore report aggregate results, compare human and automated evaluations where applicable, and analyze consistency across multiple LLM judges.

Creative reasoning systems may generate plausible but incorrect, unsupported,
or potentially misleading ideas. This risk is particularly relevant for
scientific hypothesis generation, where novel hypotheses may still lack
sufficient grounding or evidential support. We therefore evaluate grounding,
feasibility, and evidential support separately and emphasize that generated
hypotheses require independent scientific validation before practical use.

The authors are responsible for the design, analysis, interpretation, and reporting of this work and have reviewed the research for potential ethical concerns in accordance with the ICLR Code of Ethics.

\section*{Reproducibility Statement}

We provide detailed descriptions of the Universe of Thoughts algorithms, including their executable operators, scoring and selection procedures, and default hyperparameters, in Section~\ref{sec:uot_methods}, with Boden's source definitions in Appendix~\ref{app:boden-definitions} and full pseudocode, thresholds, and selection rules in Appendices~\ref{sec:uot-algorithms} and~\ref{sec:uot-scoring-details}. The experimental setup, benchmark construction, task selection, backbone models, evaluation protocols, and baseline configurations are described in Section~\ref{sec:experiments}. Evaluation prompt templates, additional implementation details, known-solution references used for novelty evaluation, and supplementary analyses are provided in the Appendix; generation-operator prompts are released with the code.

To support replication of the reported results, we provide the implementation
of UoT together with the experiment configuration files, evaluation scripts,
and prompts used in the paper as supplementary material. Where external
benchmarks are used, we follow their published evaluation protocols and
document any preprocessing or task-selection decisions made in our experiments.
Randomness, generation settings, model versions, search budgets, and other
experiment-specific parameters are reported where applicable. The novel
creative-reasoning task evaluations include both human and automated judgments,
and the HypoArena experiments additionally report judge-consistency and
operator-ablation analyses to assess the robustness of the main findings.

\bibliography{refs}

\begin{thebibliography}{41}
\providecommand{\natexlab}[1]{#1}
\providecommand{\url}[1]{\texttt{#1}}
\expandafter\ifx\csname urlstyle\endcsname\relax
  \providecommand{\doi}[1]{doi: #1}\else
  \providecommand{\doi}{doi: \begingroup \urlstyle{rm}\Url}\fi

\bibitem[Achiam et~al.(2023)Achiam, Adler, Agarwal, Ahmad, Akkaya, Aleman,
  Almeida, Altenschmidt, Altman, Anadkat, et~al.]{achiam2023gpt}
Josh Achiam, Steven Adler, Sandhini Agarwal, Lama Ahmad, Ilge Akkaya,
  Florencia~Leoni Aleman, Diogo Almeida, Janko Altenschmidt, Sam Altman,
  Shyamal Anadkat, et~al.
\newblock Gpt-4 technical report.
\newblock \emph{arXiv preprint arXiv:2303.08774}, 2023.

\bibitem[Besta et~al.(2024)Besta, Blach, Kubicek, Gerstenberger, Podstawski,
  Gianinazzi, Gajda, Lehmann, Niewiadomski, Nyczyk, et~al.]{besta2024graph}
Maciej Besta, Nils Blach, Ales Kubicek, Robert Gerstenberger, Michal
  Podstawski, Lukas Gianinazzi, Joanna Gajda, Tomasz Lehmann, Hubert
  Niewiadomski, Piotr Nyczyk, et~al.
\newblock Graph of thoughts: Solving elaborate problems with large language
  models.
\newblock In \emph{Proceedings of the AAAI conference on artificial
  intelligence}, volume~38, pp.\  17682--17690, 2024.

\bibitem[Boden(2004)]{boden2004creative}
Margaret~A Boden.
\newblock \emph{The creative mind: Myths and mechanisms}.
\newblock Routledge, 2004.

\bibitem[Boden(2007)]{boden2007creativity}
Margaret~A Boden.
\newblock Creativity in a nutshell.
\newblock \emph{Think}, 5\penalty0 (15):\penalty0 83--96, 2007.

\bibitem[Boden(2009)]{boden2009computer}
Margaret~A Boden.
\newblock Computer models of creativity.
\newblock \emph{Ai Magazine}, 30\penalty0 (3):\penalty0 23--23, 2009.

\bibitem[Chen et~al.(2025)Chen, Lin, Yuan, Zou, He, Goyal, Ong, and
  Liu]{chen2025hypospace}
Tingting Chen, Beibei Lin, Zifeng Yuan, Qiran Zou, Hongyu He, Anirudh Goyal,
  Yew-Soon Ong, and Dianbo Liu.
\newblock Hypospace: A diagnostic benchmark for set-valued hypothesis
  generation under underdetermination and sublinear coverage bounds.
\newblock \emph{arXiv preprint arXiv:2510.15614}, 2025.

\bibitem[Cheng et~al.(2024)Cheng, Yang, Jiang, Wang, Huang, Li, Li, Li, Gao,
  Li, et~al.]{cheng2024inductive}
Kewei Cheng, Jingfeng Yang, Haoming Jiang, Zhengyang Wang, Binxuan Huang,
  Ruirui Li, Shiyang Li, Zheng Li, Yifan Gao, Xian Li, et~al.
\newblock Inductive or deductive? rethinking the fundamental reasoning
  abilities of llms.
\newblock \emph{arXiv preprint arXiv:2408.00114}, 2024.

\bibitem[Chowdhery et~al.(2023)Chowdhery, Narang, Devlin, Bosma, Mishra,
  Roberts, Barham, Chung, Sutton, Gehrmann, et~al.]{chowdhery2023palm}
Aakanksha Chowdhery, Sharan Narang, Jacob Devlin, Maarten Bosma, Gaurav Mishra,
  Adam Roberts, Paul Barham, Hyung~Won Chung, Charles Sutton, Sebastian
  Gehrmann, et~al.
\newblock Palm: Scaling language modeling with pathways.
\newblock \emph{Journal of Machine Learning Research}, 24\penalty0
  (240):\penalty0 1--113, 2023.

\bibitem[Gangwal et~al.(2024)Gangwal, Ansari, Ahmad, Azad, Kumarasamy,
  Subramaniyan, and Wong]{gangwal2024generative}
Amit Gangwal, Azim Ansari, Iqrar Ahmad, Abul~Kalam Azad, Vinoth Kumarasamy,
  Vetriselvan Subramaniyan, and Ling~Shing Wong.
\newblock Generative artificial intelligence in drug discovery: basic
  framework, recent advances, challenges, and opportunities.
\newblock \emph{Frontiers in pharmacology}, 15:\penalty0 1331062, 2024.

\bibitem[Gu et~al.(2024)Gu, Wang, Zhang, and Li]{gu2024llms}
Tianyang Gu, Jingjin Wang, Zhihao Zhang, and HaoHong Li.
\newblock Llms can realize combinatorial creativity: generating creative ideas
  via llms for scientific research.
\newblock \emph{arXiv preprint arXiv:2412.14141}, 2024.

\bibitem[Guo et~al.(2025)Guo, Shariatmadari, Xiong, Huang, Kim, Williams,
  Bekiranov, and Zhang]{guo2025ideabench}
Sikun Guo, Amir~Hassan Shariatmadari, Guangzhi Xiong, Albert Huang, Myles Kim,
  Corey~M Williams, Stefan Bekiranov, and Aidong Zhang.
\newblock Ideabench: Benchmarking large language models for research idea
  generation.
\newblock In \emph{Proceedings of the 31st ACM SIGKDD Conference on Knowledge
  Discovery and Data Mining V. 2}, pp.\  5888--5899, 2025.

\bibitem[Hou et~al.(2025)Hou, Zhang, Lu, Baghel, Brei, Lu, Jiang, Brahman,
  Chaturvedi, Chang, et~al.]{hou2025creativityprism}
Zhaoyi~Joey Hou, Bowei~Alvin Zhang, Yining Lu, Bhiman~Kumar Baghel, Anneliese
  Brei, Ximing Lu, Meng Jiang, Faeze Brahman, Snigdha Chaturvedi, Haw-Shiuan
  Chang, et~al.
\newblock Creativityprism: A cross-domain evaluation framework for large
  language model creativity.
\newblock \emph{arXiv preprint arXiv:2510.20091}, 2025.

\bibitem[Lew et~al.(2026)Lew, Cao, and Buehler]{lew2026projectionbench}
Andrew~J Lew, Yuan Cao, and Markus~J Buehler.
\newblock Projectionbench: Evaluating scientific hypothesis generation in llms
  under progressive information disclosure.
\newblock \emph{arXiv preprint arXiv:2605.30284}, 2026.

\bibitem[Li et~al.(2025)Li, Zhu, Xu, Wang, and Mao]{li2025automated}
Ruizhe Li, Chiwei Zhu, Benfeng Xu, Xiaorui Wang, and Zhendong Mao.
\newblock Automated creativity evaluation for large language models: A
  reference-based approach.
\newblock In \emph{EMNLP (Findings)}, pp.\  21475--21488, 2025.

\bibitem[Liu et~al.(2026{\natexlab{a}})Liu, Zeng, and Liu]{liu2026towards}
Fan Liu, Xiaozhao Zeng, and Hao Liu.
\newblock Towards multimodal data-driven scientific discovery powered by llm
  agents.
\newblock In \emph{The Fourteenth International Conference on Learning
  Representations}, 2026{\natexlab{a}}.

\bibitem[Liu et~al.(2025)Liu, Huang, Hu, Zhou, and Tan]{liu2025hypobench}
Haokun Liu, Sicong Huang, Jingyu Hu, Yangqiaoyu Zhou, and Chenhao Tan.
\newblock Hypobench: Towards systematic and principled benchmarking for
  hypothesis generation.
\newblock \emph{arXiv preprint arXiv:2504.11524}, 2025.

\bibitem[Liu et~al.(2026{\natexlab{b}})Liu, Yang, Xie, Ni, Gao, Li, Tang,
  Ouyang, Cambria, and Zhou]{liu2026researchbench}
Yujie Liu, Zonglin Yang, Tong Xie, Jinjie Ni, Ben Gao, Yuqiang Li, Shixiang
  Tang, Wanli Ouyang, Erik Cambria, and Dongzhan Zhou.
\newblock Researchbench: Benchmarking llms in scientific discovery via
  inspiration-based task decomposition.
\newblock In \emph{Findings of the Association for Computational Linguistics:
  ACL 2026}, pp.\  13187--13207, 2026{\natexlab{b}}.

\bibitem[Lu et~al.(2024)Lu, Lu, Lange, Foerster, Clune, and Ha]{lu2024ai}
Chris Lu, Cong Lu, Robert~Tjarko Lange, Jakob Foerster, Jeff Clune, and David
  Ha.
\newblock The ai scientist: Towards fully automated open-ended scientific
  discovery.
\newblock \emph{arXiv preprint arXiv:2408.06292}, 2024.

\bibitem[Novikov et~al.(2025)Novikov, V{\~u}, Eisenberger, Dupont, Huang,
  Wagner, Shirobokov, Kozlovskii, Ruiz, Mehrabian,
  et~al.]{novikov2025alphaevolve}
Alexander Novikov, Ng{\^a}n V{\~u}, Marvin Eisenberger, Emilien Dupont, Po-Sen
  Huang, Adam~Zsolt Wagner, Sergey Shirobokov, Borislav Kozlovskii,
  Francisco~JR Ruiz, Abbas Mehrabian, et~al.
\newblock Alphaevolve: A coding agent for scientific and algorithmic discovery.
\newblock \emph{arXiv preprint arXiv:2506.13131}, 2025.

\bibitem[Panigrahi et~al.(2026)Panigrahi, Videnovi{\'c}, and
  Brbic]{panigrahi2026heurekabench}
Siba~Smarak Panigrahi, Jovana Videnovi{\'c}, and Maria Brbic.
\newblock Heurekabench: A benchmarking framework for ai co-scientist.
\newblock In \emph{International Conference on Learning Representations},
  volume 2026, pp.\  11029--11061, 2026.

\bibitem[Qiu et~al.(2025)Qiu, Zhang, Xu, Li, Song, Wang, and Zhang]{qiu2025ai}
Yansheng Qiu, Haoquan Zhang, Zhaopan Xu, Ming Li, Diping Song, Zheng Wang, and
  Kaipeng Zhang.
\newblock Ai idea bench 2025: Ai research idea generation benchmark.
\newblock \emph{arXiv preprint arXiv:2504.14191}, 2025.

\bibitem[Ruan et~al.(2026)Ruan, Wang, Hong, Wang, Liu, and
  Sun]{ruan2026evaluating}
Kai Ruan, Xuan Wang, Jixiang Hong, Peng Wang, Yang Liu, and Hao Sun.
\newblock Evaluating llms' divergent thinking capabilities for scientific idea
  generation with minimal context.
\newblock \emph{Nature communications}, 17\penalty0 (1):\penalty0 3625, 2026.

\bibitem[Runco \& Jaeger(2012)Runco and Jaeger]{runco2012standard}
Mark~A Runco and Garrett~J Jaeger.
\newblock The standard definition of creativity.
\newblock \emph{Creativity research journal}, 24\penalty0 (1):\penalty0 92--96,
  2012.

\bibitem[Schapiro et~al.(2025)Schapiro, Shashidhar, Gladstone, Black, Moon,
  Hakkani-Tur, and Varshney]{schapiro2025combinatorial}
Samuel Schapiro, Sumuk Shashidhar, Alexi Gladstone, Jonah Black, Royce Moon,
  Dilek Hakkani-Tur, and Lav~R Varshney.
\newblock Combinatorial creativity: A new frontier in generalization abilities.
\newblock \emph{arXiv preprint arXiv:2509.21043}, 2025.

\bibitem[Sen et~al.(2026)Sen, Chun, Alsagoff, Wangsajaya, Mohor, Saikia, and
  Chan]{sen2026automated}
Tan~Min Sen, Zachary Choy~Kit Chun, Syed Ali~Redha Alsagoff, Nadya~Yuki
  Wangsajaya, Banerjee Mohor, Swaagat~Bikash Saikia, and Alvin Chan.
\newblock Automated creativity evaluation of language models across open-ended
  tasks.
\newblock In \emph{Proceedings of the 64th Annual Meeting of the Association
  for Computational Linguistics (Volume 1: Long Papers)}, pp.\  23139--23173,
  2026.

\bibitem[Shin \& Kim(2025)Shin and Kim]{shin2025enhancing}
SungUk Shin and Youngjoon Kim.
\newblock Enhancing graph of thought: Enhancing prompts with llm rationales and
  dynamic temperature control.
\newblock In \emph{The Thirteenth International Conference on Learning
  Representations}, 2025.

\bibitem[Simon(1973)]{simon1973structure}
Herbert~A Simon.
\newblock The structure of ill structured problems.
\newblock \emph{Artificial intelligence}, 4\penalty0 (3-4):\penalty0 181--201,
  1973.

\bibitem[Singh et~al.(2024)Singh, Chaudhary, Singh, Soni, and
  Kapoor]{singh2024transforming}
Narotam Singh, Vaibhav Chaudhary, Nimisha Singh, Neha Soni, and Amita Kapoor.
\newblock Transforming business with generative ai: Research, innovation,
  market deployment and future shifts in business models.
\newblock \emph{arXiv preprint arXiv:2411.14437}, 2024.

\bibitem[Tian et~al.(2024)Tian, Ravichander, Qin, Le~Bras, Marjieh, Peng, Choi,
  Griffiths, and Brahman]{tian2024macgyver}
Yufei Tian, Abhilasha Ravichander, Lianhui Qin, Ronan Le~Bras, Raja Marjieh,
  Nanyun Peng, Yejin Choi, Thomas~L Griffiths, and Faeze Brahman.
\newblock Macgyver: Are large language models creative problem solvers?
\newblock In \emph{Proceedings of the 2024 Conference of the North American
  Chapter of the Association for Computational Linguistics: Human Language
  Technologies (Volume 1: Long Papers)}, pp.\  5303--5324, 2024.

\bibitem[Wang et~al.(2023)Wang, Zelikman, Poesia, Pu, Haber, and
  Goodman]{wang2023hypothesis}
Ruocheng Wang, Eric Zelikman, Gabriel Poesia, Yewen Pu, Nick Haber, and Noah~D
  Goodman.
\newblock Hypothesis search: Inductive reasoning with language models.
\newblock \emph{arXiv preprint arXiv:2309.05660}, 2023.

\bibitem[Wei et~al.(2022)Wei, Wang, Schuurmans, Bosma, Xia, Chi, Le, Zhou,
  et~al.]{wei2022chain}
Jason Wei, Xuezhi Wang, Dale Schuurmans, Maarten Bosma, Fei Xia, Ed~Chi, Quoc~V
  Le, Denny Zhou, et~al.
\newblock Chain-of-thought prompting elicits reasoning in large language
  models.
\newblock \emph{Advances in neural information processing systems},
  35:\penalty0 24824--24837, 2022.

\bibitem[Wu et~al.(2026)Wu, Fu, Si, Huang, Jiang, Li, Xia, Sun, Xu, Xiao,
  et~al.]{wu2026innovatorbench}
Yunze Wu, Dayuan Fu, Weiye Si, Zhen Huang, Mohan Jiang, Keyu Li, Shijie Xia,
  Jie Sun, Tianze Xu, Yang Xiao, et~al.
\newblock Innovatorbench: evaluating agents’ ability to conduct innovative ai
  research.
\newblock In \emph{International Conference on Learning Representations},
  volume 2026, pp.\  36144--36185, 2026.

\bibitem[Yang et~al.(2024)Yang, Liu, Gao, Xie, Li, Ouyang, Poria, Cambria, and
  Zhou]{yang2024moose}
Zonglin Yang, Wanhao Liu, Ben Gao, Tong Xie, Yuqiang Li, Wanli Ouyang, Soujanya
  Poria, Erik Cambria, and Dongzhan Zhou.
\newblock Moose-chem: Large language models for rediscovering unseen chemistry
  scientific hypotheses.
\newblock \emph{arXiv preprint arXiv:2410.07076}, 2024.

\bibitem[Yao et~al.(2023)Yao, Yu, Zhao, Shafran, Griffiths, Cao, and
  Narasimhan]{yao2023tree}
Shunyu Yao, Dian Yu, Jeffrey Zhao, Izhak Shafran, Tom Griffiths, Yuan Cao, and
  Karthik Narasimhan.
\newblock Tree of thoughts: Deliberate problem solving with large language
  models.
\newblock \emph{Advances in neural information processing systems},
  36:\penalty0 11809--11822, 2023.

\bibitem[Yasunaga et~al.(2023)Yasunaga, Chen, Li, Pasupat, Leskovec, Liang,
  Chi, and Zhou]{yasunaga2023large}
Michihiro Yasunaga, Xinyun Chen, Yujia Li, Panupong Pasupat, Jure Leskovec,
  Percy Liang, Ed~H Chi, and Denny Zhou.
\newblock Large language models as analogical reasoners.
\newblock \emph{arXiv preprint arXiv:2310.01714}, 2023.

\bibitem[Yu et~al.(2023)Yu, He, and Ying]{yu2023thought}
Junchi Yu, Ran He, and Rex Ying.
\newblock Thought propagation: An analogical approach to complex reasoning with
  large language models.
\newblock \emph{arXiv preprint arXiv:2310.03965}, 2023.

\bibitem[Zhang et~al.(2026)Zhang, Xu, Zheng, Yu, Zhu, Luo, Wei, Qiao, Du,
  Zheng, et~al.]{zhang2026innogym}
Jintian Zhang, Kewei Xu, Jingsheng Zheng, Zhuoyun Yu, Yuqi Zhu, Yujie Luo,
  Lanning Wei, Shuofei Qiao, Lun Du, Da~Zheng, et~al.
\newblock Innogym: Benchmarking the innovation potential of ai agents.
\newblock In \emph{International Conference on Learning Representations},
  volume 2026, pp.\  71349--71379, 2026.

\bibitem[Zheng et~al.(2026)Zheng, Tam, Nguyen, Xu, Wang, Jiayang, Tsang, Wang,
  Bai, Fang, et~al.]{zheng2026newtonbench}
Tianshi Zheng, Kiu~Wai Tam, Kim Hue~Nam Nguyen, Baixuan Xu, Zhaowei Wang, Cheng
  Jiayang, Hong~Ting Tsang, Weiqi Wang, Jiaxin Bai, Tianqing Fang, et~al.
\newblock Newtonbench: Benchmarking generalizable scientific law discovery in
  llm agents.
\newblock In \emph{International Conference on Learning Representations},
  volume 2026, pp.\  99018--99088, 2026.

\bibitem[Zhong et~al.(2026)Zhong, Jiang, Wang, Chen, Lu, Ye, Shi, Yang, Wang,
  Li, et~al.]{zhong2026before}
Tianyun Zhong, Wangyi Jiang, Wei Wang, Xuanang Chen, Yaojie Lu, Shiwei Ye,
  Yuzhen Shi, Boyu Yang, Jinghang Wang, Han Li, et~al.
\newblock Before the action: Benchmarking llms on prospective hypothesis
  discovery.
\newblock \emph{arXiv preprint arXiv:2607.15766}, 2026.

\bibitem[Zhou et~al.(2023)Zhou, Yan, Shlapentokh-Rothman, Wang, and
  Wang]{zhou2023language}
Andy Zhou, Kai Yan, Michal Shlapentokh-Rothman, Haohan Wang, and Yu-Xiong Wang.
\newblock Language agent tree search unifies reasoning acting and planning in
  language models.
\newblock \emph{arXiv preprint arXiv:2310.04406}, 2023.

\bibitem[Zhou et~al.(2024)Zhou, Pujara, Ren, Chen, Cheng, Le, Chi, Zhou,
  Mishra, and Zheng]{zhou2024self}
Pei Zhou, Jay Pujara, Xiang Ren, Xinyun Chen, Heng-Tze Cheng, Quoc~V Le,
  Ed~Chi, Denny Zhou, Swaroop Mishra, and Huaixiu~Steven Zheng.
\newblock Self-discover: Large language models self-compose reasoning
  structures.
\newblock \emph{Advances in Neural Information Processing Systems},
  37:\penalty0 126032--126058, 2024.

\end{thebibliography}
\bibliographystyle{iclr2027_conference}

% Flush bibliography floats and start the appendix on a fresh page so appendix
% tables cannot migrate backward into the reference list.
\clearpage
\appendix
\section{Connection to Boden's Definitions}
\label{app:boden-definitions}
\label{sec:methodology-details}

Our computational formulation follows Boden's distinction among
combinational, exploratory, and transformational creativity
\citep{boden2004creative,boden2007creativity,boden2009computer}.
Boden characterizes the three modes as follows
(bracketed glosses are added by us):

\begin{enumerate}
    \item \textbf{Combinational creativity.}
    ``Combinational creativity, produces unfamiliar combinations
    [i.e., solutions] of familiar ideas, and it works by making
    associations between ideas that were previously only indirectly
    linked.''

    \item \textbf{Exploratory creativity.}
    ``In exploratory creativity, the person moves through the
    [solution] space, exploring it to find out what's there
    (including previously unvisited locations [i.e., new ideas])---and,
    in the most interesting cases, to discover both the potential and
    the limits of the space in question.''

    \item \textbf{Transformational creativity.}
    ``In transformational creativity, a [solution] space or style
    itself is transformed by altering (or dropping) one or more of its
    defining dimensions [i.e., rules].''
\end{enumerate}

We translate these cognitive descriptions into computational search
operations. In C-UoT, an ``unfamiliar combination'' is operationalized as
transferring a functionally relevant mechanism from a contextually different
source into the target problem. In E-UoT, movement through the existing
conceptual space is operationalized as expanding the currently explored
candidate set while keeping the problem rules fixed. In T-UoT, transformation
of the conceptual space is operationalized as modifying mutable rules or
assumptions while preserving the task's immutable requirements.

\begin{table}[H]
\caption{Correspondence between Boden's creativity modes and UoT operators.}
\label{tab:boden-uot-map}
\centering
\small
\begin{tabular}{p{0.32\textwidth} p{0.30\textwidth} p{0.28\textwidth}}
\toprule
\bf Boden concept & \bf Computational interpretation & \bf UoT operator \\
\midrule
Unfamiliar combinations of familiar ideas
  & Import structure from another context
  & analogue $\rightarrow$ mechanism $\rightarrow$ transfer \\
Explore previously unvisited locations in an existing space
  & Expand the candidate set with rules fixed
  & \textsc{Shift} / \textsc{Deepen} \\
Alter defining dimensions of the space
  & Modify mutable assumptions or rules
  & gated rule transformation \\
\bottomrule
\end{tabular}
\end{table}

We reinterpret these three modes as distinct strategies for a common
computational problem:

\begin{quote}
\emph{Given a problem and the solutions already explored, what should a
reasoning system change in order to reach a useful solution that has not yet
been considered?}
\end{quote}

This perspective yields three different search regimes. Combinational
reasoning changes the \emph{source of structure} used to construct a solution;
exploratory reasoning changes the \emph{candidate solution} while preserving
the current rules; and transformational reasoning changes the
\emph{rules or assumptions} that define which solutions are admissible.
These distinctions define the computational structure of Universe of Thoughts
(UoT). The staged conditional-generation formalization of each mode appears in
Section~\ref{sec:uot_methods}; full pseudocode and selection details are
provided in Appendix~\ref{sec:uot-algorithms} and
Appendix~\ref{sec:uot-scoring-details}.

\section{UoT Pseudocode}
\label{sec:uot-algorithms}

This appendix specifies the executable operator interfaces and the control flow
for Combinational, Exploratory, and Transformational UoT
(Section~\ref{sec:uot_methods}). The operators name what is generated or
evaluated; benchmark-specific prompts instantiate the corresponding
representation (policy components for novel creative-reasoning tasks; mechanisms and
hypotheses for HypoArena). Soft-gate thresholds default to
$\tau_g{=}\tau_i{=}2$.

\subsection{Executable Operator Interfaces}
\label{app:uot-operator-interfaces}

All UoT operators are instantiated as structured LLM calls. Across both
evaluation settings, the three modes share the same search objects---external
structure, candidate solutions under fixed rules, and mutable
assumptions---while differing in intermediate representation and selection
budgets.

\paragraph{C-UoT.}
\textsc{GenerateAnalogues} proposes source problems that are functionally
related to the target but contextually distinct from already explored
solutions. \textsc{SolveSource} obtains a source-side solution or structured
account of how the analogue addresses its problem.
\textsc{ExtractMechanisms} removes source-specific surface detail and retains
the transferable mechanism. \textsc{TransferAndSynthesize} instantiates that
mechanism in the target problem and produces a candidate solution or
hypothesis. Soft quality filters and final ranking then retain promising
transfers (Appendix~\ref{sec:uot-scoring-details}).

\paragraph{E-UoT.}
\textsc{Expand} generates neighboring candidates while preserving the problem
formulation and mutable rules. Each expansion explores either a different
direction by changing the current mechanism (\textsc{Shift}) or an
underexplored consequence of the current mechanism (\textsc{Deepen}).
Candidates are evaluated and retained by the shared search controller
($\mathrm{LLM}_{\mathrm{vote}}$ / beam retention), matching the Tree-of-Thoughts
control structure used by the ToT baseline. \textsc{Finalize} converts the
final beam into submitted outputs.

\paragraph{T-UoT.}
\textsc{ExtractRules} identifies mutable assumptions and separates them from
immutable task requirements. \textsc{GenerateTransformations} proposes
modifications to those assumptions (e.g., remove, relax, invert, add, or
replace). A retained transformation must preserve the original task goal and
core difficulty while opening a meaningfully different solution region.
\textsc{Reframe} constructs the working problem representation under the
selected transformed rules, after which \textsc{CreativeSearch} runs
downstream solution search in that reframed space.

\subsection{Algorithms}

\begin{algorithm}[H]
\caption{Combinational UoT (HypoArena: $R_{\max}{=}5$ with soft gate;
novel tasks: final F/U/N ranking)}
\label{alg:cuot}
\KwIn{Target problem $P_t$; known/explored solutions $\mathcal{K}_t$}
\KwOut{Ranked candidates $\mathcal{Y}_C$}
$\mathcal{Y} \leftarrow \emptyset$\;
\For{$r=1,\ldots,R_{\max}$}{
    $\mathcal{A} \leftarrow \mathrm{LLM}_{\mathrm{propose}}(\textsc{GenerateAnalogues}\mid P_t,\mathcal{K}_t\cup\mathcal{Y};\ \text{far})$\;
    \ForEach{$a \in \mathcal{A}$}{
        $\mathcal{S}_a \leftarrow \mathrm{LLM}_{\mathrm{propose}}(\textsc{SolveSource}\mid a)$\;
        $\mathcal{Z}_a \leftarrow \mathrm{LLM}_{\mathrm{propose}}(\textsc{ExtractMechanisms}\mid a,\mathcal{S}_a)$\;
        $y \leftarrow \mathrm{LLM}_{\mathrm{propose}}(\textsc{TransferAndSynthesize}\mid P_t,\mathcal{Z}_a)$\;
        $(g,i) \leftarrow \mathrm{LLM}_{\mathrm{score}}(y\mid P_t;\ \text{grounding/insight}\in\{1..5\})$\;
        \tcp*[h]{HypoArena soft gate; novel creative-reasoning tasks defer filtering to final F/U/N ranking}
        \If{$g\geq\tau_g \wedge i\geq\tau_i$}{
            $\mathcal{Y} \leftarrow \mathcal{Y}\cup\{y\}$\;
        }
    }
}
\ForEach{$y \in \mathcal{Y}$}{
    $\mathrm{Score}(y) \leftarrow \mathrm{LLM}_{\mathrm{score}}(y\mid P_t,\mathcal{K}_t;\ \text{pair axes or F/U/N})$\;
}
$\mathcal{Y}_C \leftarrow \mathrm{TopK}(\mathcal{Y};\, \mathrm{Score})$\;
\Return{$\mathcal{Y}_C$}\;
\end{algorithm}

\begin{algorithm}[H]
\caption{Exploratory UoT (defaults: $T{=}4$, $k{=}5$, $b{=}|\mathcal{Y}_E|$)}
\label{alg:euot}
\KwIn{Problem $P$; initial beam $\mathcal{B}^{(0)}$; depth $T$; per-node budget $k$; beam width $b$}
\KwOut{Finalized candidates $\mathcal{Y}_E$}
\For{$t=0,\ldots,T-1$}{
    $\widetilde{\mathcal{B}}^{(t+1)} \leftarrow \emptyset$\;
    \ForEach{$s \in \mathcal{B}^{(t)}$}{
        $\mathcal{C} \leftarrow \mathrm{LLM}_{\mathrm{propose}}(\textsc{Expand}\mid P,s;\, |\mathcal{C}|{\le}k)$\;
        \tcp*[h]{Each $c$ carries $\mathrm{op}(c)\in\{\textsc{Shift},\textsc{Deepen}\}$}
        $\widetilde{\mathcal{B}}^{(t+1)} \leftarrow \widetilde{\mathcal{B}}^{(t+1)} \cup \mathcal{C}$\;
    }
    $\mathcal{I} \leftarrow \mathrm{LLM}_{\mathrm{vote}}(\widetilde{\mathcal{B}}^{(t+1)};\, |\mathcal{I}|{=}b)$\;
    $\mathcal{B}^{(t+1)} \leftarrow \{\widetilde{\mathcal{B}}^{(t+1)}_i : i\in\mathcal{I}\}$\;
}
$\mathcal{Y}_E \leftarrow \mathrm{LLM}_{\mathrm{propose}}(\textsc{Finalize}\mid P,\mathcal{B}^{(T)})$\;
\Return{$\mathcal{Y}_E$}\;
\end{algorithm}

\begin{algorithm}[H]
\caption{Transformational UoT (paper: $k{=}5$; $S_{\mathrm{select}}$ as below)}
\label{alg:tuot}
\KwIn{Problem $P$; immutable constraints $H$}
\KwOut{Candidates $\mathcal{Y}_T$}
$R \leftarrow \mathrm{LLM}_{\mathrm{propose}}(\textsc{ExtractRules}\mid P,H)$\;
$\mathcal{R}' \leftarrow \mathrm{LLM}_{\mathrm{propose}}(\textsc{GenerateTransformations}\mid P,H,R;\, |\mathcal{R}'|{=}k)$\;
$\mathcal{R}^{*} \leftarrow \emptyset$\;
\ForEach{$R' \in \mathcal{R}'$}{
    $(b_{\mathrm{goal}}, b_{\mathrm{diff}}, b_{\mathrm{space}})
    \leftarrow \mathrm{LLM}_{\mathrm{score}}(P,H,R,R';\ \{0,1\}^3)$\;
    \tcp*[h]{$b_{\mathrm{goal}}$: goal preserved; $b_{\mathrm{diff}}$: difficulty preserved; $b_{\mathrm{space}}$: solution space opened}
    \If{$b_{\mathrm{goal}} \wedge b_{\mathrm{diff}} \wedge b_{\mathrm{space}}$}{
        $\mathcal{R}^{*} \leftarrow \mathcal{R}^{*} \cup \{R'\}$\;
    }
}
$R^{*} \leftarrow S_{\mathrm{select}}(\mathcal{R}^{*})$\;
$P' \leftarrow \textsc{Reframe}(P,R^{*})$\;
$\mathcal{Y}_T \leftarrow \textsc{CreativeSearch}(P')$\;
\Return{$\mathcal{Y}_T$}\;
\end{algorithm}

\FloatBarrier

\section{Scoring and Selection Protocol}
\label{sec:uot-scoring-details}

Throughout UoT,
$\mathrm{LLM}_{\mathrm{propose}}$,
$\mathrm{LLM}_{\mathrm{score}}$, and
$\mathrm{LLM}_{\mathrm{vote}}$
denote structured backbone calls that return candidate lists, task-specific
scores, or discrete selection indices. Lightweight deterministic filters are
applied where specified.

These interfaces implement the operational criteria of each search regime:
functional relevance and unfamiliarity for combinational transfer, candidate
selection during exploratory expansion, and goal/difficulty preservation
together with solution-space opening for transformational search. Exact
generation-operator prompts are released with the code; Appendix~\ref{sec:evaluation-metrics-appendix}
provides the novel creative-reasoning task evaluation prompts used for scoring.

\subsection{Novel Creative-Reasoning Task Configuration}
\label{sec:scoring-controlled}

\paragraph{C-UoT.}
Final candidates are ranked by feasibility, utility, and novelty for a total
score of $0$--$100$: feasibility and utility contribute $0$--$30$ each and
novelty contributes $0$--$40$. Defaults are $10$ analogues, $5$ solutions per
analogue, and $20$ combinational candidates before TopK selection. Soft
grounding/insight gates used in HypoArena are deferred; filtering is performed
by the final F/U/N ranking.

\paragraph{E-UoT.}
E-UoT does not use a continuous creativity score inside the search loop.
$\mathrm{LLM}_{\mathrm{vote}}$ instead returns exactly $b$ candidate indices,
using the same voting interface as the matched ToT baseline. The default
configuration is $T=4$, $k=5$, and $b=|\mathcal{Y}_E|$.

\paragraph{T-UoT.}
Each transformation is evaluated using the Boolean gate
$(b_{\mathrm{goal}},b_{\mathrm{diff}},b_{\mathrm{space}})$.
$S_{\mathrm{select}}$ is set-valued over gated transforms, and
\textsc{CreativeSearch} runs E-UoT under each retained transform.

\subsection{HypoArena Configuration}
\label{sec:scoring-hypoarena}

\paragraph{C-UoT.}
Each transferred candidate is evaluated during search on
$(g,i)\in\{1,\ldots,5\}^2$, corresponding to grounding and insight.
Candidates are retained when $g\geq2$ and $i\geq2$, with at most
$R_{\max}=5$ proposal rounds and typically three analogues per target tension.

Final ranking uses the HypoArena-pair criteria on a $0$--$100$ scale:
grounding, insight, justification, and redundancy. Grounding acts as a hard
gate with threshold $\tau=50$. Among candidates passing the gate, ranking is

$$
\mathrm{mean}(g,i,j)
-
\lambda
\left(
\frac{\mathrm{redundancy}}{100}
\right),
\qquad
\lambda=1.5.
$$

Mechanism-diverse TopK selection produces the final submitted set.

\paragraph{E-UoT.}
The exploratory controller matches the novel creative-reasoning task defaults ($T=4$,
$k=5$, $b=|\mathcal{Y}_E|$), with $\mathrm{LLM}_{\mathrm{vote}}$ returning
exactly $b$ indices as in the matched ToT baseline.

\paragraph{T-UoT.}
Each transformation is evaluated with the same Boolean gate
$(b_{\mathrm{goal}},b_{\mathrm{diff}},b_{\mathrm{space}})$.
$S_{\mathrm{select}}$ follows the paper configuration: an analogue-probe
score argmax in which one analogue is generated under each surviving
transformed observation, its structural richness is scored programmatically,
and

$$
R^*
=
\arg\max_{R'}
\mathrm{ProbeScore}(R').
$$

The default alternative is an LLM index vote over the gated set.
\textsc{CreativeSearch} is Alg.~\ref{alg:cuot} on the reframed observation.

\section{Supplementary Results}

\subsection{Novel Creative-Reasoning Tasks}
\label{sec:controlled-tasks}

Table~\ref{tasks} gives the full objective, context, and constraint specifications
for the three novel creative-reasoning tasks summarized in Section~\ref{sec:experiments}.

\begin{table*}[t]
\caption{Novel creative-reasoning tasks}
\label{tasks}
\centering
\begin{tabular}{l p{0.75\textwidth}}
\hline
\multicolumn{1}{c}{\bf Task} & \multicolumn{1}{c}{\bf Description} \\
\hline

\textit{Task 1: One-Lane Bridge} &
\textbf{Objective:} Design a traffic-control policy that minimizes average vehicle delay. \newline
\textbf{Context:} A single-lane bridge with known traffic patterns, fixed crossing times, and direction-switching penalties. \newline
\textbf{Constraints:} No new infrastructure; traffic must remain unidirectional at all times. \\
\hline

\textit{Task 2: Electricity Tariff} &
\textbf{Objective:} Design a tariff and demand-response program that reduces daily peak load. \newline
\textbf{Context:} A residential feeder experiences predictable periods of high demand. \newline
\textbf{Constraints:} No new physical capacity; critical medical loads must never be curtailed. \\
\hline

\textit{Task 3: Social Cohesion} &
\textbf{Objective:} Design an intervention that strengthens cross-group cohesion, measured by new ties, mixing, and mutual aid. \newline
\textbf{Context:} A community contains distinct social groups and a fixed intervention period. \newline
\textbf{Constraints:} The intervention must protect privacy, remain inclusive, and maintain a safe environment. \\
\hline

\end{tabular}
\end{table*}

\FloatBarrier

\subsection{Novel Creative-Reasoning Task Full Results}
\label{sec:controlled-full-results}

We generate 10 independent solutions for each method and report the mean score
across generations. Tables~\ref{human-results-combined} and~\ref{all-results-combined}
report the full Novelty, Utility, and Feasibility decomposition corresponding
to the Creativity summary in Table~\ref{tab:controlled-creativity}.
Creativity is the mean of Novelty and Utility.

As contextual reference points, GPT-5 and DeepSeek V3.1 are evaluated under a
single task prompt, allowing each model to rely on its own internal reasoning
capabilities. These are not matched backbone comparisons. GPT-4o-based T-UoT
achieves higher Creativity than GPT-5 in four of six task--evaluator
comparisons and than DeepSeek V3.1 in three of six; the advantage is
concentrated on Bridge and Electricity, while the frontier models lead on the
more ambiguous Society task.

\begin{table*}[t]
\caption{Human evaluation results across all creative reasoning tasks.
C=Creativity, N=Novelty, U=Utility, and F=Feasibility. Bold: column best;
underline: column second-best.}
\label{human-results-combined}
\centering
\scriptsize
\begin{tabular}{lcccccccccccc}
\toprule
\multicolumn{1}{c}{\bf Method} & \multicolumn{4}{c}{\bf Bridge Task} & \multicolumn{4}{c}{\bf Electricity Task} & \multicolumn{4}{c}{\bf Society Task} \\
\cmidrule(lr){2-5} \cmidrule(lr){6-9} \cmidrule(lr){10-13}
 & \bf C & \bf N & \bf U & \bf F & \bf C & \bf N & \bf U & \bf F & \bf C & \bf N & \bf U & \bf F \\
\midrule
OnePrompt & 0.480 & 0.450 & 0.500 & 0.820 & 0.540 & 0.480 & \underline{0.610} & 0.870 & 0.550 & 0.490 & 0.610 & \textbf{0.810} \\
CoT & 0.590 & 0.610 & 0.570 & 0.750 & 0.560 & 0.520 & 0.600 & \textbf{0.920} & 0.500 & 0.430 & 0.580 & \textbf{0.810} \\
ToT & 0.580 & 0.600 & 0.560 & 0.680 & 0.530 & 0.540 & 0.520 & 0.760 & 0.530 & 0.440 & 0.610 & \textbf{0.810} \\
EGoT & \underline{0.650} & \underline{0.650} & \underline{0.650} & 0.810 & 0.500 & 0.440 & 0.550 & \underline{0.880} & 0.560 & 0.480 & \underline{0.640} & 0.800 \\
C-UoT & 0.580 & 0.550 & 0.610 & 0.820 & 0.440 & 0.440 & 0.440 & 0.870 & 0.560 & 0.500 & 0.620 & \textbf{0.810} \\
E-UoT & 0.600 & 0.600 & 0.610 & 0.790 & 0.430 & 0.380 & 0.470 & 0.740 & 0.550 & 0.480 & 0.620 & 0.730 \\
T-UoT & \textbf{0.680} & \textbf{0.690} & \textbf{0.670} & 0.770 & \textbf{0.620} & \textbf{0.610} & \textbf{0.640} & 0.800 & 0.550 & 0.490 & 0.600 & 0.730 \\
GPT-5 & 0.620 & 0.610 & 0.640 & \underline{0.840} & \underline{0.580} & \underline{0.600} & 0.560 & 0.770 & \underline{0.640} & \underline{0.680} & 0.600 & 0.770 \\
DeepSeek V3.1 & 0.460 & 0.460 & 0.470 & \textbf{0.910} & \underline{0.580} & 0.580 & 0.580 & 0.750 & \textbf{0.720} & \textbf{0.690} & \textbf{0.740} & \textbf{0.810} \\
\bottomrule
\end{tabular}
\end{table*}

\begin{table*}[t]
\caption{LLM evaluation results across all creative reasoning tasks.
C=Creativity, N=Novelty, U=Utility, and F=Feasibility. Bold: column best;
underline: column second-best.}
\label{all-results-combined}
\centering
\scriptsize
\begin{tabular}{lcccccccccccc}
\toprule
\multicolumn{1}{c}{\bf Method} & \multicolumn{4}{c}{\bf Bridge Task} & \multicolumn{4}{c}{\bf Electricity Task} & \multicolumn{4}{c}{\bf Society Task} \\
\cmidrule(lr){2-5} \cmidrule(lr){6-9} \cmidrule(lr){10-13}
 & \bf C & \bf N & \bf U & \bf F & \bf C & \bf N & \bf U & \bf F & \bf C & \bf N & \bf U & \bf F \\
\midrule
OnePrompt & 0.566 & 0.732 & 0.400 & 0.925 & 0.508 & 0.587 & 0.430 & 0.680 & 0.522 & 0.675 & 0.370 & 0.730 \\
CoT & 0.651 & 0.742 & 0.560 & 0.980 & 0.450 & 0.541 & 0.360 & 0.675 & 0.586 & 0.732 & 0.440 & 0.560 \\
ToT & 0.579 & \underline{0.799} & 0.360 & 0.900 & 0.488 & \textbf{0.657} & 0.320 & 0.610 & 0.574 & 0.727 & 0.420 & 0.495 \\
EGoT & 0.641 & 0.643 & \underline{0.640} & \textbf{1.000} & 0.409 & 0.428 & 0.390 & \textbf{0.825} & 0.613 & 0.775 & 0.450 & 0.525 \\
C-UoT & \underline{0.664} & 0.728 & 0.600 & \textbf{1.000} & 0.512 & 0.563 & 0.460 & 0.675 & \underline{0.649} & 0.799 & 0.500 & 0.530 \\
E-UoT & 0.589 & 0.778 & 0.400 & 1.000 & 0.463 & 0.587 & 0.340 & 0.755 & 0.545 & \underline{0.810} & 0.280 & 0.575 \\
T-UoT & \textbf{0.698} & \textbf{0.807} & 0.590 & 0.930 & \underline{0.526} & \underline{0.641} & 0.410 & 0.665 & 0.628 & \textbf{0.846} & 0.410 & 0.575 \\
GPT-5 & 0.649 & 0.649 & \textbf{0.650} & \textbf{1.000} & 0.504 & 0.497 & \underline{0.510} & \underline{0.815} & 0.631 & 0.652 & \underline{0.610} & \textbf{0.870} \\
DeepSeek V3.1 & 0.458 & 0.476 & 0.440 & 0.950 & \textbf{0.571} & 0.563 & \textbf{0.580} & 0.770 & \textbf{0.723} & 0.767 & \textbf{0.680} & \underline{0.825} \\
\bottomrule
\end{tabular}
\end{table*}

\FloatBarrier

\subsection{Human--LLM Judge Agreement}
\label{sec:human-llm-agreement}

To assess the reliability of automated evaluation, we compare Gemini~2.5~Pro
scores with human judgments. At the method level ($n=27$ method--task cells),
Creativity shows strong agreement ($r=0.708$, $p<10^{-4}$), while Utility shows
moderate agreement ($r=0.481$). Agreement is weaker for Novelty and
Feasibility. Instance-level correlations are lower, suggesting that the LLM
judge is more reliable for comparing aggregate method behavior than for
assigning scores to individual solutions.

\begin{table}[t]
\caption{Pearson correlation between human and Gemini~2.5~Pro scores on the novel creative-reasoning tasks (identical Utility/Feasibility prompts). Method level: one cell per method$\times$task ($n{=}27$). Instance level: paired individual solutions ($n{=}267$). Bold: strongest correlation in each row block.}
\label{tab:human-llm-corr}
\centering
\small
\begin{tabular}{l ccc}
\toprule
\bf Metric & \bf $r$ & \bf $p$ & \bf $n$ \\
\midrule
\multicolumn{4}{l}{\emph{Method level}} \\
Creativity & \textbf{0.708} & $3.6{\times}10^{-5}$ & 27 \\
Utility    & 0.481          & $1.1{\times}10^{-2}$ & 27 \\
Novelty    & 0.224          & $2.6{\times}10^{-1}$ & 27 \\
Feasibility& 0.039          & $8.5{\times}10^{-1}$ & 27 \\
\midrule
\multicolumn{4}{l}{\emph{Instance level}} \\
Creativity & \textbf{0.342} & $9.6{\times}10^{-9}$ & 267 \\
Utility    & 0.249          & $3.9{\times}10^{-5}$ & 267 \\
Novelty    & 0.168          & $5.8{\times}10^{-3}$ & 267 \\
Feasibility& 0.067          & $2.7{\times}10^{-1}$ & 267 \\
\bottomrule
\end{tabular}
\end{table}

\subsection{Sample Solutions for Bridge Task}
\label{app:bridge-examples}

Sample solutions for the Bridge Task are provided in Table~\ref{sample-bridge} for illustration.

\begin{table}[t]
\caption{Sample solutions for the Bridge task}
\label{sample-bridge}
\begin{center}
\begin{tabular}{lp{12cm}}
\multicolumn{1}{c}{\bf Method}  &\multicolumn{1}{c}{\bf Solution}
\\ \hline \\
C-UoT         &The core functional mechanism is a cooperative negotiation system. Its essential components are a driver interface for signaling crossing intent (input) and a central allocation system. The system aggregates these individual intents to form a comprehensive demand profile, and based on this, allocates specific crossing times (output) back to the drivers. This interaction, inspired by negotiation protocols in distributed systems, ensures fair and efficient resource use. \\
E-UoT             &Based on a feedback-based mechanism, this solution implements a dynamic scheduling model that uses real-time traffic data (input) to continuously adjust vehicle crossing schedules. This interaction dynamically allocates crossing windows based on current conditions (output), which serves to optimize overall flow and reduce vehicle idle times. \\
T-UoT            &This solution models the crossing scenario using game theory, treating vehicles as rational players aiming to minimize their individual travel time. The system analyzes the strategic interactions between all vehicles and calculates a Nash Equilibrium---a state where no single vehicle can benefit by unilaterally changing its crossing strategy. The resulting equilibrium strategy is then implemented as the global crossing schedule, ensuring a stable and efficient outcome for the entire system. \\
\end{tabular}
\end{center}
\end{table}

\subsection{Novel Creative-Reasoning Task Sensitivity Analysis}
\label{sec:controlled-sensitivity}

\begin{figure}[ht]
\centering
% Subfigure for the number of analogous problems
\begin{subfigure}[b]{0.49\columnwidth}
    \centering
    \includegraphics[height=2in]{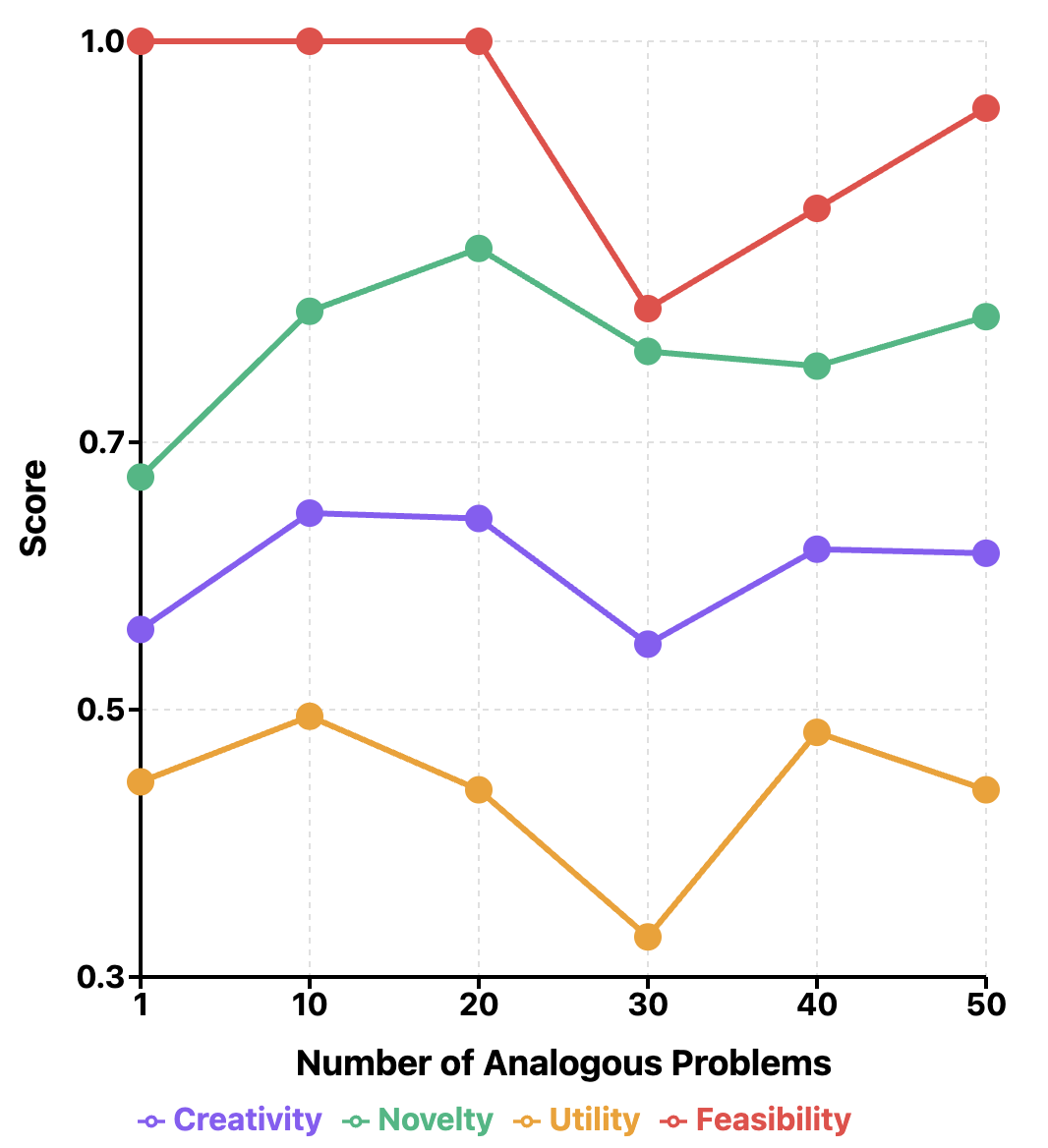}
    \caption{Effect of number of analogous problems}
    \label{fig:num_analogous}
\end{subfigure}
\hfill % This creates a space between the two subfigures
% Subfigure for the number of solutions per problem
\begin{subfigure}[b]{0.49\columnwidth}
    \centering
    \includegraphics[height=2in]{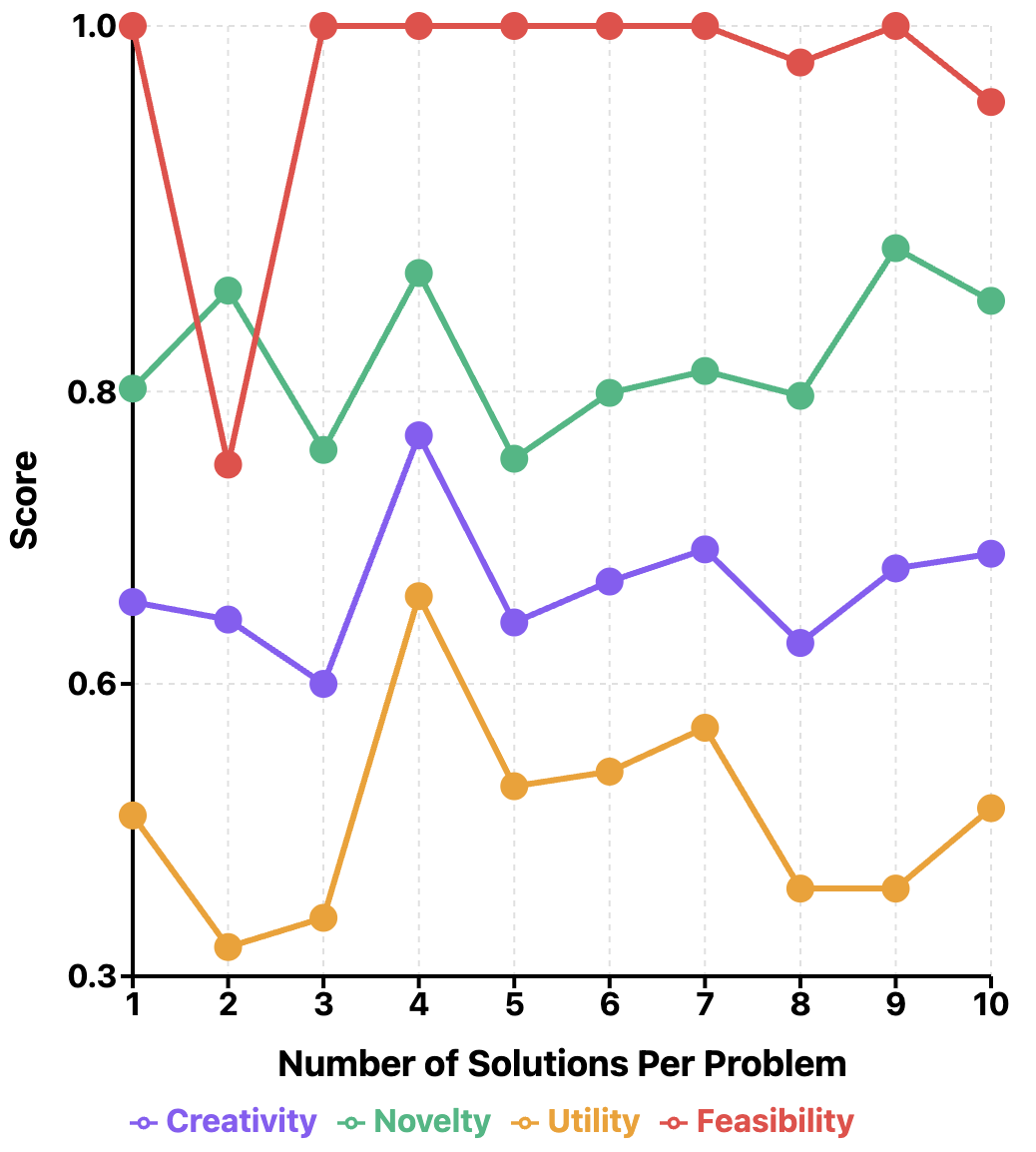}
    \caption{Effect of number of solutions generated}
    \label{fig:num_solution}
\end{subfigure}
\caption{\textit{Sensitivity analysis for UoT}}
\label{fig:sensitivity_analysis}
\end{figure}

Figures \ref{fig:num_analogous} and \ref{fig:num_solution} present sensitivity of C-UoT's performance on the Bridge Task with respect to two key parameters: the number of analogous problems and the number of solutions generated per problem.

Figure \ref{fig:num_analogous} shows that performance follows a U-shaped curve as the number of analogous problems increases. We hypothesize this indicates three distinct regimes: a small number of analogies may yield a few high-quality, easily combinable thoughts; a moderate number can introduce conceptual noise that hinders effective synthesis; and a very large number provides a sufficiently rich and diverse pool of concepts where the benefits of flexibility overcome the noise.

Figure \ref{fig:num_solution} shows that increasing the number of solutions per analogous problem, while exhibiting some fluctuation, generally leads to improved performance. This suggests that a larger pool of ``conceptual primitives'' or thoughts derived from each analogy provides more raw material for synthesizing a high-quality final solution.

\FloatBarrier

\subsection{GLM-4-9B HypoArena Results}
\label{sec:hypoarena-glm-results}

Table~\ref{tab:hypoarena-glm-100} reports the same 100-task HypoArena panel
with GLM-4-9B as the generation backbone. Unlike the Qwen3-14B setting
(Table~\ref{tab:hypoarena-qwen-100}), OnePrompt ranks first overall
(Avg.\ BTD $1560.4$; WR $46.3\%$), while E-UoT remains the strongest
structured search method (Avg.\ BTD $1542.1$; WR $40.7\%$), ahead of ToT,
CoT, EGoT, C-UoT, and T-UoT. This reversal relative to Qwen indicates that
which creativity mode is most effective can depend on backbone capability,
even when the task panel and judge are held fixed.

\begin{table}[t]
\caption{GLM-4-9B on HypoArena (100 tasks; Gemma-3-12B judge; canonicalized core hypotheses). Bio\,34 / ML\,34 / Social\,32. Bio/ML/Social/Avg: pooled BTD ratings centered at 1500. WR: pairwise win rate (\%). Bold: column best.}
\label{tab:hypoarena-glm-100}
\centering
\small
\begin{tabular}{r l ccccc}
\toprule
\bf Rank & \bf Method & \bf Bio & \bf ML & \bf Social & \bf Avg & \bf WR (\%) \\
\midrule
1 & OnePrompt & \textbf{1577.9} & 1537.0          & \textbf{1564.6} & \textbf{1560.4} & \textbf{46.3} \\
2 & E-UoT     & 1558.8          & \textbf{1545.4} & 1522.4          & 1542.1          & 40.7 \\
3 & ToT       & 1504.0          & 1528.2          & 1516.2          & 1516.1          & 34.0 \\
4 & CoT       & 1506.2          & 1475.3          & 1518.1          & 1500.0          & 30.2 \\
5 & EGoT      & 1438.8          & 1520.3          & 1442.1          & 1467.1          & 25.2 \\
6 & C-UoT     & 1444.6          & 1454.9          & 1483.5          & 1460.7          & 21.7 \\
7 & T-UoT     & 1469.7          & 1439.0          & 1453.0          & 1453.6          & 19.0 \\
\bottomrule
\end{tabular}
\end{table}

\subsection{Judge Consistency on Frozen Qwen-100 Hypotheses}
\label{sec:judge-consistency}

To assess whether the HypoArena results depend on a particular LLM judge, we
re-score the same frozen Qwen3-14B hypotheses with Mistral-Small-24B and
Prometheus-2 while holding all generated outputs fixed. Table~\ref{tab:judge-consistency-pairs}
reports agreement in pairwise outcomes, hard preference reversals, and Spearman
correlation over case--method win shares. Table~\ref{tab:judge-consistency-contrasts}
further measures consistency in the relative advantage of UoT variants over
key baselines on a per-case basis.

Across judge pairs, case--method rankings show moderate agreement
($\rho=0.587$--$0.616$), while hard A$\leftrightarrow$B reversals remain rare
($2.6$--$5.4\%$). More importantly, the relative advantage of E-UoT over
OnePrompt and ToT is consistently preserved across judges
($\rho=0.59$--$0.79$), with similarly moderate consistency for C-UoT and
T-UoT relative to ToT. These results indicate that exact scores vary across
judges, but the main comparative trends are not driven by a single evaluator.

\begin{table}[t]
\caption{Judge consistency on frozen Qwen-100 hypotheses (Gemma-3-12B, Mistral-Small-24B, Prometheus-2). Spearman $\rho$ over case$\times$method win-shares ($n{=}700$). Exact 3-way agreement and hard A$\leftrightarrow$B flip over $n{=}2100$ matched pairs. Bold: column best (lowest flip).}
\label{tab:judge-consistency-pairs}
\centering
\small
\begin{tabular}{l ccc}
\toprule
\bf Judge pair & \bf Spearman $\rho$ & \bf Exact 3-way & \bf Hard A$\leftrightarrow$B flip \\
\midrule
Gemma $\leftrightarrow$ Mistral     & \textbf{0.616} & \textbf{0.604} & \textbf{2.6\%} \\
Gemma $\leftrightarrow$ Prometheus  & 0.587          & 0.478          & 4.4\% \\
Mistral $\leftrightarrow$ Prometheus & 0.604          & 0.520          & 5.4\% \\
\bottomrule
\end{tabular}
\end{table}

\begin{table}[t]
\caption{Spearman $\rho$ on per-case method contrasts across judges (G$=$Gemma, M$=$Mistral, P$=$Prometheus). Bold: row best.}
\label{tab:judge-consistency-contrasts}
\centering
\small
\begin{tabular}{l ccc}
\toprule
\bf Contrast & \bf G$\leftrightarrow$M & \bf G$\leftrightarrow$P & \bf M$\leftrightarrow$P \\
\midrule
$\Delta$(E $-$ OnePrompt) & 0.71          & 0.71          & \textbf{0.79} \\
$\Delta$(E $-$ ToT)       & 0.65          & 0.59          & \textbf{0.69} \\
$\Delta$(C $-$ ToT)       & \textbf{0.67} & 0.49          & 0.52 \\
$\Delta$(T $-$ ToT)       & \textbf{0.63} & 0.51          & 0.57 \\
\bottomrule
\end{tabular}
\end{table}

\FloatBarrier

\subsection{Qualitative HypoArena Examples and Non-Redundant Successes}
\label{sec:uot-qualitative}
\label{sec:hypoarena-qualitative}

Table~\ref{tab:uot-qualitative} presents illustrative high-performing cases
for each UoT mode. For visualization, we select one case per mode in which
that mode ranks first among the seven methods with a substantial margin over
the next-best method. These examples are not used as evidence of aggregate
performance; rather, they illustrate the qualitatively different reasoning
operations induced by combinational, exploratory, and transformational UoT.
We compare the canonicalized core hypothesis produced by each UoT mode with
the ToT hypothesis from the same frozen Qwen3-14B generations.

\begin{table*}[t]
\caption{Illustrative high-performing HypoArena cases for each UoT mode
(Qwen3-14B; Gemma-3-12B judge). Each row shows a case in which the listed
mode ranks first among seven methods with a substantial margin over the
next-best method. Hypotheses are abbreviated to their canonicalized core
claims.}
\label{tab:uot-qualitative}
\centering
\scriptsize
\begin{tabular}{p{0.07\textwidth} p{0.08\textwidth} p{0.08\textwidth} p{0.20\textwidth} p{0.24\textwidth} p{0.22\textwidth}}
\toprule
\bf Mode & \bf Domain & \bf Gap to \#2 & \bf Task (brief) & \bf UoT idea & \bf ToT idea \\
\midrule
C-UoT
  & ML
  & $+0.33$ vs E-UoT
  & Why entropy-minimization test-time adaptation fails for vision--language models under shift
  & Standard Shannon entropy assumes static class uniformity and clashes with pretraining-induced confidence asymmetries; use class-specific Tsallis entropy shapes tuned to those biases
  & Reweight entropy with class-specific ``niche factors'' so adaptation favors views that keep the correct class despite imperfect top-1 scores
  \\
\midrule
E-UoT
  & Social
  & $+0.27$ vs EGoT
  & What sustains adherence to digital fitness coaches (live, avatar, or agent)
  & Retention depends on \emph{dynamic} alignment: anthropomorphic design plus real-time shifts in emotional vs.\ technical support as user needs and adherence phases change
  & Engagement is a static balance of instrumental utility and para-social bonds, mediated by anthropomorphism, authenticity, and support type
  \\
\midrule
T-UoT
  & Social
  & $+0.15$ vs OnePrompt
  & Why holiday e-commerce returns spike beyond ordinary product dissatisfaction
  & Surges arise when shoppers privilege local cues (gift urgency, immediate satisfaction) over systemic factors (return windows, recipient mismatch, delayed evaluation)---heuristic misalignment, not quality failure
  & Returns are driven by social-identity pressure and temporal dissonance between purchase intent and post-holiday reevaluation
  \\
\bottomrule
\end{tabular}
\end{table*}

The examples illustrate distinct search behaviors. C-UoT introduces structure
from an external analogy, E-UoT develops a new hypothesis while retaining the
original conceptual framing, and T-UoT changes the explanatory assumptions
that define the hypothesis space. These qualitative differences complement
the aggregate results by showing that the UoT variants do not merely
produce alternative phrasings of similar hypotheses, but induce different
forms of creative search.

\paragraph{Non-redundant C-UoT and T-UoT successes.}
Although C-UoT and T-UoT rank below ToT on average, they exhibit
nontrivial pockets of strong performance
(Table~\ref{tab:ct-success-pockets}). Under the primary Gemma judge,
C-UoT is top-ranked on 15/100 tasks (8 uniquely), while T-UoT is
top-ranked on 11/100 tasks (6 uniquely). These task-specific advantages
also recur across evaluators: C-UoT and T-UoT outperform ToT on 26 and
24 tasks, respectively, under at least two of three judges, and on 11
tasks each under all three judges. Thus, their weaker aggregate rankings
do not imply that the corresponding search modes are uniformly
unproductive; rather, they appear to access useful hypothesis directions
on a subset of problems.

\begin{table}[t]
\caption{Task-level success frequency on Qwen3-14B HypoArena.
Top-ranked counts use the primary Gemma judge; cross-judge columns
report tasks on which the method outperforms ToT under multiple judges.}
\label{tab:ct-success-pockets}
\centering
\small
\begin{tabular}{lcccc}
\toprule
Method & Top-1 & Sole Top-1 & $>\,$ToT, $\geq$2/3 judges & $>\,$ToT, 3/3 judges \\
\midrule
C-UoT & 15 & 8 & 26 & 11 \\
T-UoT & 11 & 6 & 24 & 11 \\
\bottomrule
\end{tabular}
\end{table}

\FloatBarrier

\subsection{Failure-Mode Diagnosis and Interventions}
\label{sec:uot-intervention}

\paragraph{Stage-level diagnosis.}
The C-UoT trace analysis suggests that failures often occur after analogy
retrieval rather than because no meaningful analogy is found
(Table~\ref{tab:cuot-stage-diagnosis}). An analogue and mapping are selected
on essentially all tasks, and mapped-mechanism specificity is nearly identical
when C-UoT does and does not beat ToT. In contrast, preservation of mapped
mechanism content into the final hypothesis is higher when C-UoT beats ToT,
with strong transfer more frequent in wins than non-wins. This pattern
motivates the downstream-synthesis intervention below.

For T-UoT, transformation execution alone does not distinguish success
(Table~\ref{tab:tuot-stage-diagnosis}): rule changes are applied in both
winning and losing cases, and lexical preservation into the final hypothesis
is weak in both groups. Among the 12 largest applied-transform failures,
7/12 involved reasonable transformations that were not converted into
competitive hypotheses, while 5/12 involved weak, misaimed, or derailing
transformations. This suggests a dual bottleneck for T-UoT: downstream
consequence reasoning is often weak, while the transformation gate itself
can also admit poor rule changes.

To test whether the diagnosed C-UoT and T-UoT bottlenecks are actionable, we
perform targeted interventions on frozen failure cases while keeping the
upstream creative search fixed (Table~\ref{tab:uot-intervention}). For C-UoT,
we preserve the retrieved analogy and mapping and modify only the final
synthesis using a human-written rewrite; this intervention tests whether the
frozen intermediate analogy can support a stronger final hypothesis when
downstream synthesis is improved. For T-UoT, we keep the selected
transformation fixed but require the model to explicitly derive one
mechanistic consequence of that transformation before producing the final
hypothesis.

\begin{samepage}
We report
\[
\Delta = P(\mathrm{intervened}\succ\mathrm{original}) - 0.5,
\]
averaged over four random seeds, together with the resulting rank among the
intervened method and five fixed baselines. Positive $\Delta$ indicates that
correcting the diagnosed downstream stage improves the final hypothesis without
changing the upstream creative operation.
\end{samepage}

\FloatBarrier

\begin{table}[H]
\caption{C-UoT stage-level diagnosis relative to ToT on Qwen-100.
Wins denote tasks where C-UoT is preferred to ToT by the primary judge.}
\label{tab:cuot-stage-diagnosis}
\centering
\small
\begin{tabular}{lcc}
\toprule
Signal & C-UoT $>$ ToT & C-UoT $\not>$ ToT \\
\midrule
Analogy selected & 100\% & 100\% \\
Mapped-mechanism specificity & 0.97 & 0.97 \\
Mapped $\rightarrow$ final preservation & 0.53 & 0.43 \\
Strong transfer & 52\% & 32\% \\
\bottomrule
\end{tabular}
\end{table}

\begin{table}[H]
\caption{T-UoT stage-level diagnosis relative to ToT on Qwen-100.}
\label{tab:tuot-stage-diagnosis}
\centering
\small
\begin{tabular}{lcc}
\toprule
Signal & T-UoT $>$ ToT & T-UoT $\not>$ ToT \\
\midrule
Gate passed & 100\% & 100\% \\
Transform applied & 75\% & 83\% \\
Rules $\rightarrow$ framing & 0.75 & 0.83 \\
Rules $\rightarrow$ final & 0.13 & 0.16 \\
\bottomrule
\end{tabular}
\end{table}

\begin{table}[H]
\caption{Targeted failure-mode interventions on frozen C-UoT and T-UoT cases.
$\Delta$ denotes $P(\mathrm{intervened}\succ\mathrm{original})-0.5$ averaged over four seeds.
Rank is computed among the intervened method and five fixed baselines; lower is better.}
\label{tab:uot-intervention}
\centering
\scriptsize
\setlength{\tabcolsep}{3pt}
\begin{tabular}{p{0.06\textwidth} p{0.07\textwidth} p{0.16\textwidth} p{0.14\textwidth} p{0.07\textwidth} p{0.08\textwidth} p{0.28\textwidth}}
\toprule
\bf Mode & \bf Domain & \bf Task summary & \bf Intervened step & \bf Mean $\Delta$ & \bf Rank & \bf What changed (before $\to$ after) \\
\midrule
C-UoT
  & Social
  & Why do vending-machine choices become habitually locked onto familiar snacks despite healthier alternatives?
  & Final synthesis only (mapping frozen). Human rewrites the final from the frozen positive-feedback / path-dependence mapping; no new analogy search.
  & $+0.28$
  & $\#4\to\#1.5$ (\#1 on 3/4 seeds)
  & \textbf{Before:} generic ``habit via cues/placement'' without the mapped feedback loop or a discriminating test.
  \textbf{After:} keeps the frozen local positive-feedback story (repeated choice reinforces favorites and suppresses alternatives) and adds the mapped prediction---disrupting cues should raise switching, whereas merely placing healthier items near favorites should not.
  Mapping-token preserve rises $0.20\to0.33$.
  \\
\midrule
T-UoT
  & Bio
  & Why are ponerine ants much richer in some regions---diversification-rate differences, occupancy time, or dispersal source--sink structure?
  & Post-transform synthesis. Transform rules stay frozen (relax fossil/dispersal assumptions); model must state one mechanistic consequence of those rules, then write the final.
  & $+0.25$
  & $\#7\to\#2.5$
  & \textbf{Before:} retreats to data-limitation language (incomplete fossils, uneven sampling, barriers) and never uses the transform.
  \textbf{After:} applies the consequence of the relaxed rules---dispersal/colonization \emph{sequence} imposes geographic constraints, so richness tracks early vs.\ late colonization timing rather than present-day diversification rates.
  \\
\bottomrule
\end{tabular}
\end{table}

Both interventions improve the corresponding failure cases. For C-UoT,
preserving and explicitly realizing the frozen analogy mapping raises the
hypothesis from an average rank of \#4 to \#1.5. For T-UoT, forcing the model
to derive a mechanistic consequence of the selected transformation improves
the average rank from \#7 to \#2.5. These interventions are diagnostic rather
than complete repairs, but they show that the creative operation itself can
contain useful structure that is lost during downstream synthesis. This
supports our failure analysis: a key limitation of C-UoT and T-UoT is not
always failure to discover an analogy or transformation, but failure to
faithfully integrate that creative move into the final grounded hypothesis.

\FloatBarrier

\subsection{Inference Cost versus Quality}
\label{sec:hypoarena-token-cost}

A natural concern is that UoT gains might simply reflect larger inference
budgets. Table~\ref{tab:hypoarena-token-cost} reports mean generation tokens
per task on a fixed 10-task cost-audit subset (4 biomedical / 3 machine
learning / 3 social science; Qwen3-14B), alongside the full 100-task BTD
ratings and win rates from Table~\ref{tab:hypoarena-qwen-100}. Agentic methods
are substantially more expensive than OnePrompt/CoT, as expected. However,
quality does not increase monotonically with tokens: E-UoT achieves the best
Avg.\ BTD (1535.4) at $10.0\times$ OnePrompt cost
($\approx$1.35$\times$ ToT), whereas EGoT, C-UoT, and T-UoT require
$27.6$--$52.5\times$ OnePrompt cost ($2.8$--$5.3\times$ E-UoT) yet achieve lower
BTD than E-UoT. Thus, on this backbone, search
structure---not token count alone---appears to drive the HypoArena ranking.

\begin{table}[t]
\caption{Inference cost vs.\ HypoArena quality (Qwen3-14B). Tokens/task: mean
total tokens on a fixed 10-task cost-audit subset (4 Bio / 3 ML / 3 Social).
Relative: vs.\ OnePrompt on the same subset. Avg.\ BTD and WR (\%): full
100-task panel (Table~\ref{tab:hypoarena-qwen-100}). Bold: column best
(lowest cost / highest quality).}
\label{tab:hypoarena-token-cost}
\centering
\small
\begin{tabular}{l r r r r}
\toprule
\bf Method & \bf Tokens/task & \bf Rel.\ cost & \bf Avg.\ BTD & \bf WR (\%) \\
\midrule
OnePrompt & 3.7k          & \textbf{1.0}$\times$ & 1475.7 & 14.3 \\
CoT       & \textbf{3.5k} & 0.9$\times$          & 1493.7 & 18.8 \\
ToT       & 27.1k         & 7.4$\times$          & 1515.8 & 25.5 \\
EGoT      & 101.6k        & 27.6$\times$         & 1505.1 & 25.3 \\
C-UoT     & 134.3k        & 36.5$\times$         & 1489.6 & 19.5 \\
E-UoT     & 36.7k         & 10.0$\times$         & \textbf{1535.4} & \textbf{32.7} \\
T-UoT     & 193.3k        & 52.5$\times$         & 1484.6 & 17.3 \\
\bottomrule
\end{tabular}
\end{table}

\FloatBarrier

\section{Experimental Design}
\label{sec:experimental-design}

\subsection{Novel Creative-Reasoning Task Construction and Validation}
\label{sec:controlled-task-construction}

Although the novel creative-reasoning tasks do not rely on external domain simulators
(e.g., physics engines), they were constructed and validated through a
human-in-the-loop process to ensure logical consistency and accessibility.
We first generated ten candidate open-ended tasks, then manually solved each
candidate to verify that the problem statement was coherent and that the
solution space was reachable. From this pool we selected three diverse tasks
(One-Lane Bridge, Electricity Tariff, and Social Cohesion) and refined their
rules to balance assessable constraints against sufficient open-endedness for
creative reasoning.

Our goal was to isolate creative reasoning rather than domain-expertise
retrieval. Heavy domain-specific constraints that require complex engineering
simulations would risk confounding the two. By restricting the tasks to
self-contained logic settings validated by human solvers, solution validity is
defined by explicit task constraints and objectives, which supports
reproducible evaluation.

\subsection{Benchmark Selection Rationale}
\label{sec:benchmark-rationale}

As discussed in Section~\ref{sec:benchmarking-creativity}, existing creativity
benchmarks tend to emphasize either highly open-ended generation or more
tightly constrained problem solving. Open-ended ideation tasks permit diverse
outputs but often make functional success difficult to assess objectively,
whereas benchmarks such as MacGyver \citep{tian2024macgyver} and scientific
hypothesis-generation tasks provide clearer problem definitions but impose
stronger constraints on the admissible solution space. Our novel creative-reasoning tasks and
HypoArena evaluation are chosen as complementary settings that jointly cover
clear objectives with expansive solution spaces and scientifically grounded
multi-hypothesis comparison.

\subsection{HypoArena Protocol}
\label{sec:hypoarena-protocol}

HypoArena provides scientific hypothesis-generation cases grounded in source
documents. We restrict evaluation to its three single-hypothesis research
domains---biomedical science, machine learning, and social science. The
remaining domains---financial analysis, IT operations, and safety
investigation---require multi-hypothesis submissions and are therefore outside
the scope of our single-hypothesis generation setting. From the eligible
research-domain pool, we evaluate a fixed panel of 100 tasks, stratified
approximately evenly across domains (34 biomedical, 34 machine learning, and
32 social science). The same panel is held fixed across all methods and
backbone variants.

Because different reasoning methods can produce outputs with substantially
different levels of detail, we first canonicalize each final response into its
main core hypothesis before pairwise evaluation. This ensures that comparisons
focus on the underlying scientific idea rather than rewarding longer or more
elaborated explanations.

HypoArena is not designed as a direct metric of creativity; its primary rubric
dimensions emphasize Contextual Grounding and Evidential Justification. At the
same time, the Inferential Insight dimension explicitly asks whether a
hypothesis synthesizes context observations into a \emph{non-obvious} but still
context-constrained explanatory or mechanistic claim. Because judges therefore
partially reward non-obvious scientific insight when it remains well
supported, creative hypotheses can still receive high scores.

For each task, an LLM judge compares every pair of methods using bidirectional
prompts and a five-level preference scale ranging from a clear preference for
one hypothesis to a clear preference for the other, with ties permitted.
Judgments follow the HypoArena rubric, emphasizing context grounding and
analytical quality rather than stylistic fluency or verbosity.

We use Gemma-3-12B as the primary judge. Pairwise outcomes are aggregated with
the Bradley--Terry--Davidson (BTD) model, initialized at 1500, and we
additionally report pairwise win rate. BTD ratings are computed separately
within each domain, and the overall score is the mean of the biomedical,
machine-learning, and social-science ratings. Judge-consistency analyses under
additional evaluators are reported in Appendix~\ref{sec:judge-consistency}.

\subsection{Baselines}
\label{sec:baselines-details}

Our primary goal is to evaluate whether explicitly modeling creative reasoning
provides benefits beyond general-purpose reasoning and search strategies. We
compare UoT against the following established reasoning frameworks across both
evaluation settings:

\begin{itemize}[nosep]
    \item \textit{OnePrompt}: The task description is provided directly to the model without an explicit reasoning scaffold.
    \item \textit{Chain-of-Thought (CoT)} \citep{wei2022chain}: The model generates a sequential step-by-step reasoning trajectory.
    \item \textit{Tree of Thoughts (ToT)} \citep{yao2023tree}: The model explores multiple branching reasoning trajectories through iterative generation, evaluation, and selection.
    \item \textit{Enhanced Graph of Thoughts (EGoT)} \citep{shin2025enhancing}: A graph-based reasoning framework that supports generation, aggregation, and refinement of intermediate thoughts.
\end{itemize}

These baselines span increasingly structured forms of general-purpose
reasoning, from direct generation to sequential, tree-based, and graph-based
search. This allows us to test whether explicitly modeling creative reasoning
provides benefits beyond increasing the complexity of generic reasoning search.

We intentionally focus on general-purpose reasoning frameworks rather than
domain-specific scientific-discovery agents. UoT is designed as a
domain-agnostic creative reasoning framework, and our HypoArena experiment
tests whether the same creative reasoning procedures transfer to scientific
hypothesis generation under matched reasoning conditions, rather than whether
UoT outperforms specialized discovery systems that rely on additional retrieval
pipelines, tools, or domain-specific assumptions.

\subsection{Implementation Details}
\label{sec:implementation-details}

\paragraph{Novel creative-reasoning tasks.}
We use GPT-4o as the backbone for UoT and all reasoning baselines, holding the
underlying model fixed so that comparisons more directly reflect differences
among reasoning frameworks. We additionally report GPT-5 and DeepSeek V3.1 as
contextual reference models; these are not treated as matched baselines because
their internal reasoning procedures and computational budgets are not directly
comparable.

\paragraph{HypoArena.}
Our primary experiments use Qwen3-14B as the shared backbone for UoT and all
baseline reasoning methods. We use the same task inputs, backbone, and shared
decoding settings where applicable. To assess whether the observed behavior
depends on a particular backbone, we additionally evaluate selected settings
with GLM-4-9B. Unless otherwise stated, HypoArena results refer to the
Qwen3-14B experiments. We use open-weight models for HypoArena rather than
GPT-4o because the 100-task, multi-method setup requires a large number of
generations and pairwise judgments, making proprietary API evaluation
prohibitively expensive under our budget.

\section{Novel Creative-Reasoning Task Evaluation Metrics and Prompts}
\label{sec:evaluation-metrics-appendix}

Our three-task evaluation framework is guided by the classic definition of creativity \citep{runco2012standard}, which comprises two primary components: Utility and Novelty. We also introduce Feasibility as a critical third metric to ensure solutions are valid and practical by adhering to all problem constraints.

A key challenge in evaluating LLM-generated outputs is their varying granularity. One model might offer a concise core idea, while another provides a detailed plan with sub-solutions. To ensure a fair comparison focused on the generated core creative insight, we introduce a Solution Canonicalization step. Before scoring, each generated solution is processed by an LLM (Gemini~2.5~Pro, temperature~$0$) to extract its single, core conceptual idea. This pre-processing step standardizes all outputs to a consistent format, enabling an equitable assessment of creativity. Following canonicalization, each solution is assessed using the metrics below:
\begin{itemize}[nosep]
    \item \textit{Utility}: This metric measures how effectively the solution solves the problem. It is scored on a continuous scale from 0 to 1. An LLM evaluator places the solution on a $0$--$10$ rubric relative to known solutions (and an optional optimal-solution upper reference), then rescales the score to $[0,1]$.
    \item \textit{Novelty}: To measure conceptual novelty, we first extract a short conceptual fingerprint (``theme'') of the canonicalized solution with Gemini~2.5~Pro. We then embed that theme with \texttt{gemini-embedding-001} and compute its cosine distance from the embeddings of known solution concepts. The novelty score is $1 - \max(\mathrm{similarity})$ to known concepts, normalized by the maximum known-to-optimal concept distance so that the optimal reference provides a practical upper bound on the scale.
    \item \textit{Feasibility}: This metric is scored from 0 to 1 and assesses whether a solution is plausible and meets the rules of the problem. The LLM evaluator is provided with a pre-defined checklist of all explicit constraints and scores the solution based on adherence to those constraints.
\end{itemize}

The final \textit{Creativity Score} is the mean of the Utility and Novelty scores. Feasibility is reported separately as a validity check.

\subsection{Known Solutions for Novelty and Utility}
\label{sec:known-solutions}

For each novel creative-reasoning task, we maintain a fixed set of \emph{known solutions}: author-curated conventional baselines that represent standard approaches to the problem, not outputs of any evaluated model. Novelty compares each candidate against short conceptual fingerprints of these baselines (\texttt{known\_solutions\_concept}); Utility uses the corresponding fuller descriptions as calibration anchors. The baselines used in our experiments are:

\paragraph{Bridge.}
(1)~Fixed time blocks (alternate direction on a timer).
(2)~Queue-based switching (switch when the active queue is empty).
(3)~Batch processing (serve a fixed batch size before switching).

\paragraph{Electricity.}
(1)~Time-of-use pricing with static peak/off-peak blocks.
(2)~Critical peak pricing on declared stress events.
(3)~Peak-time rebates for voluntary load reduction.

\paragraph{Society.}
(1)~Structured mixers that force cross-group interaction.
(2)~Cross-group micro-grants requiring multi-group teams.
(3)~Randomized one-on-one ``coffee chat'' pairing.
(4)~A shared resource library that creates interdependence.

Human and automated evaluations use the same Utility and Feasibility prompts. Automated evaluation uses Gemini~2.5~Pro with temperature set to $0$. The prompts below are the templates used in our implementation; placeholders such as \texttt{[SOLUTION]}, \texttt{[TASK\_DESCRIPTION]}, \texttt{[KNOWN\_SOLUTIONS]}, \texttt{[OPTIMAL\_SOLUTION]}, and \texttt{[CHECK\_POINTS]} are filled per task and candidate.

\subsection{Canonicalization Prompt}
\begin{verbatim}
You are a solution processing assistant. Your task is to identify
all individual solutions from the text provided and canonicalize
each one into a concise paragraph that functions as an operational
blueprint.

Text to Process:
[SOLUTION]

Instructions:
1. Scan the text and identify the boundaries of each distinct
   solution.
2. For each solution, author a self-contained paragraph describing
   its core functional mechanism. This paragraph must identify the
   essential components, their primary inputs and outputs, and
   explain how they interact to produce the final result.
3. If an inspirational domain is mentioned, include it for context.
4. The level of detail must be sufficient to make the process
   unambiguous and reproducible in principle, but should not be a
   granular, step-by-step implementation guide. Omit low-level
   details like specific parameters or mathematical formulas.

Return your response as a single, valid JSON array of strings,
where each string is a descriptive paragraph.
\end{verbatim}

\subsection{Feasibility Prompt}
\begin{verbatim}
Evaluate the feasibility of the following solution against these
check points:

Solution: [SOLUTION]

Check Points:
[CHECK_POINTS]

Return a JSON object with two fields:
- "score": a number between 0.0 and 1.0 representing the
  feasibility score, where:
  - 0.0 = None of the check points are met
  - 1.0 = All check points are fully met
  - Values in between represent partial fulfillment
- "reasoning": a 1-2 sentence explanation of your evaluation

Please score it on the assumption that the underlying technology
and strategy are sound.
\end{verbatim}

\subsection{Utility Prompt}
\begin{verbatim}
You are an expert systems evaluator. Your task is to assess the
utility of the "Current Solution" by placing it on a percentile
scale relative to the provided "Known Solutions."

Optimal Solution Reference (Score 10 - Maximum Utility):
[OPTIMAL_SOLUTION]

Task Description:
[TASK_DESCRIPTION]

Known Solutions (Consider these a representative sample of
standard, conventional approaches):
[KNOWN_SOLUTIONS]

Current Solution to Evaluate:
[SOLUTION]

Instructions:
1. Analyze the Current Solution's effectiveness at achieving the
   primary objective in the Task Description.
2. Compare this effectiveness to the quality of the Known Solutions.
3. Place the Current Solution on a scale from 0 to 10 based on the
   rubric below. A score of 5 means it is a minor, incremental
   improvement over the average known solution.

Evaluate the effectiveness of the core idea itself, not the quality
of its description or implementation plan.

Scoring Rubric (Absolute Improvement Scale):
* 0-2 (Incremental): A minor or trivial improvement over the
  baseline. The core strategy is still conventional and offers only
  marginal gains.
* 3-5 (Competent): A solid, well-defined solution that is clearly
  more effective than the baseline. It represents a competent, but
  not necessarily innovative, approach.
* 6-8 (Significant): A significant conceptual or practical
  improvement. The strategy is advanced, highly effective, and
  demonstrates a sophisticated understanding of the problem's
  dynamics.
* 9-10 (Transformative): A state-of-the-art or paradigm-shifting
  approach. The solution reframes the problem or introduces a
  mechanism that is fundamentally superior to conventional
  strategies.

Return a single JSON object with two fields:
- "score": A single integer from 0 to 10 representing the placement.
- "reasoning": A 1-2 sentence justification for the score,
  explicitly referencing the rubric and comparing the Current
  Solution to the Known Solutions.
\end{verbatim}

\subsection{Novelty Theme-Extraction Prompt}
\begin{verbatim}
Deconstruct the following solution into its core operational logic
to create a "conceptual fingerprint." Your goal is to create a
single, dense phrase that describes the solution's fundamental
system of operation, ignoring prose and justifications.

To do this, you will:
1. Identify the Core Mechanism: The central process or tool
   (e.g., a timer, an auction, a sensor).
2. Identify the Actors & Interaction: The key components and what
   they do (e.g., vehicles submit bids).
3. Identify the Governing Principle: The underlying rule that
   dictates the outcome (e.g., fixed intervals, willingness to pay).
4. Synthesize: Combine these elements into a single, logical phrase
   that describes the system.

Crucial Rule: Do NOT summarize into high-level keywords. The output
must describe the unique process or system rule.

Solution to Analyze:
[SOLUTION]

Return format:
"A system where [actors] [interaction] based on the principle of
[governing principle]."
\end{verbatim}

After theme extraction, Novelty is computed with
\texttt{gemini-embedding-001} embeddings as described above
(cosine distance to known concepts, normalized by the known-to-optimal
concept distance).

\end{document}